\documentclass{article}
\usepackage{CJKutf8}

\usepackage{graphicx}
\graphicspath{{./}{figures/}}  

\usepackage{subcaption}
\usepackage{adjustbox}
\usepackage{float}
\usepackage{grffile}
\usepackage{amsmath}
\usepackage{amssymb}
\usepackage{authblk}
\usepackage[sorting=none]{biblatex}
\usepackage{csquotes}
\usepackage{algorithm}
\usepackage{algpseudocode}
\usepackage{booktabs}
\usepackage{multirow}
\usepackage{makecell}
\usepackage{array}

\newcommand{\ImgH}{5cm}

\usepackage[english]{babel}
\usepackage[hidelinks]{hyperref}

\title{Intelligent Image Search Algorithms Fusing Visual Large Models}

\author[1,\textdagger]{Kehan Wang} 
\author[2,\textdagger]{Tingqiong Cui} 
\author[2]{Yang Zhang} 
\author[2]{Yu Chen} 
\author[3,*]{Shifeng Wu} 
\author[3,*]{Zhenzhang Li} 
\affil[1]{Chongqing University, Chongqing China} 
\affil[2]{CRRC Chongqing Co., Ltd., Chongqing China} 
\affil[3]{Guangdong Polytechnic Normal University, Guangzhou China} 
\affil[$\dagger$]{\small These authors contributed equally to this work.}
\affil[*]{\small Corresponding authors}
\date{November 2025}

\begin{document}

\maketitle

\noindent \textbf{Email Correspondence:} \\
Kehan Wang (Co-first): \href{mailto:7696718@qq.com}{7696718@qq.com} \\
Tingqiong Cui (Co-first): \href{mailto:cuitingqiong.cqgs@crrcgc.cc}{cuitingqiong.cqgs@crrcgc.cc} \\
Shifeng Wu (Corresponding): \href{mailto:fengtree@126.com}{fengtree@126.com} \\
Zhenzhang Li (Corresponding): \href{mailto:zhenzhangli@gpnu.edu.cn}{zhenzhangli@gpnu.edu.cn} \\
Yibu Yang: \href{mailto:423044337@qq.com}{423044337@qq.com} \\
Yang Zhang: \href{mailto:1037442999@qq.com}{1037442999@qq.com} \\
Yu Chen: \href{mailto:zhangyang19840520@163.com}{zhangyang19840520@163.com}

\clearpage
\begin{abstract}
    Fine-grained image retrieval, which aims to find images containing specific object components and assess their detailed states, is critical in fields like security and industrial inspection. However, conventional methods face significant limitations: manual features (e.g., SIFT) lack robustness; deep learning-based detectors (e.g., YOLO) can identify component presence but cannot perform state-specific retrieval or zero-shot search; Visual Large Models (VLMs) offer semantic and zero-shot capabilities but suffer from poor spatial grounding and high computational cost, making them inefficient for direct retrieval. To bridge these gaps, this paper proposes DetVLM, a novel intelligent image search framework that synergistically fuses object detection with VLMs. The framework pioneers a search-enhancement paradigm via a two-stage pipeline: a YOLO detector first conducts efficient, high-recall component-level screening to determine component presence; then, a VLM acts as a recall-enhancement unit, performing secondary verification for components missed by the detector. This architecture directly enables two advanced capabilities: 1) State Search: Guided by task-specific prompts, the VLM refines results by verifying component existence and executing sophisticated state judgments (e.g., "sun visor lowered"), allowing retrieval based on component state. 2) Zero-shot Search: The framework leverages the VLM's inherent zero-shot capability to recognize and retrieve images containing unseen components or attributes (e.g., "driver wearing a mask") without any task-specific training. Experiments on a vehicle component dataset show DetVLM achieves a state-of-the-art overall retrieval accuracy of 94.82\%, significantly outperforming detection-only baselines. It also attains 94.95\% accuracy in zero-shot search for driver mask-wearing and over 90\% average accuracy in state search tasks. DetVLM provides a robust and practical solution for fine-grained image retrieval that seamlessly integrates basic, state-based, and zero-shot search.  
    
\end{abstract}

\section{Introduction}

Fine-grained image retrieval, defined by the core objective of retrieving images that contain specific object components and further targeting those with predefined semantic states, stands as a pivotal and challenging frontier in computer vision\cite{krause20133d, 7780803,6909576,ge2019deepfashion2}. Its practical value lies in translating real-world user demands into accurate visual search results--demands spanning diverse high-stakes fields such as railway transportation security, where operators face urgent needs for train appearance-centric retrieval. This includes finding images of trains with specific exterior features or abnormal appearance states such as peeled paint, damaged exterior panels, or open emergency doors. Such tasks are critical for operational scheduling and safety monitoring but are severely hindered by real-world interferences including motion blur from high-speed operation, drastic variations in illumination between tunnels and open lines, and frequent occlusions that obscure key appearance details. Beyond the railway sector, in traffic surveillance and public security, law enforcement agencies require the capability to retrieve images of vehicles where certain components are present and in a specific state, such as a masked driver or a lowered sun visor, from massive volumes of monitoring footage, as manual review is profoundly inefficient and prone to human error\cite{2016A,7553002,Sultani2018RealWorldAD}. Similarly, in industrial quality control, manufacturers may need to retrieve images of workpieces with defective components, such as cracked engine parts—a scenario that relies on precise component-centric search and often incorporates state attributes. Despite its broad and critical applicability, achieving a retrieval system that is simultaneously robust, efficient, and semantically aware remains an open and formidable challenge\cite{ge2019deepfashion2,8954181}. 

A systematic analysis of the current technological landscape reveals that conventional approaches to fine-grained retrieval are characterized by fundamental and complementary limitations, creating a significant performance bottleneck. Existing systems are predominantly architected around a single modality, forcing a suboptimal choice between efficiency and semantic understanding. Traditional hand-crafted feature methods, exemplified by SIFT and HOG\cite{lowe2004distinctive,navneet2011histograms,herbert2006surf}, fundamentally lack the robustness to cope with real-world variations such as viewpoint changes, significant lighting differences, or motion blur, rendering them unable to stably match fine-grained appearance features across diverse and unpredictable scenarios. Deep learning-based object detectors, particularly the advanced iterations of the YOLO series\cite{2016You,2017YOLO9000,2018YOLOv3,Bochkovskiy2020YOLOv4OS,Jocher_Ultralytics_YOLO_2023,yolo11_ultralytics,tian2025yolov12}, have made remarkable strides in efficiently determining the presence of components within images, enabling them to quickly identify whether a target component exists in an image\cite{7410526,7485869,2016SSD}. However, these models are inherently designed for "existence confirmation" and coarse categorization; they are semantically blind, possessing no inherent capability to retrieve objects based on nuanced appearance features or to understand and query semantic states, nor can they generalize to zero-shot scenarios for recognizing a new component type entirely absent from the training data\cite{9879567,2025Grounding,10.1007/978-3-031-20080-9_42}. Conversely, Visual Large Models (VLMs)\cite{kim2021vilt,radford2021learning,Jia2021ScalingUV,li2022january,Li2023BLIP2BL,NEURIPS2021_50525975}, which are pre-trained on vast corpora of image-text pairs, possess formidable semantic reasoning and zero-shot recognition capabilities, allowing them to understand natural language descriptions of appearance features and states. Nevertheless, VLMs deployed directly for retrieval are insufficiently accurate due to their inherent lack of precise spatial understanding, often misattributing features to incorrect regions or missing small but critical details\cite{NEURIPS2021_50525975,li2022january,zhong2022regionclip,9879567,2025Grounding}. Furthermore, their high computational intensity makes them impractical for processing large-scale datasets, such as thousands of hours of  railway surveillance footage, in an efficient manner\cite{kim2021vilt,radford2021learning,Jia2021ScalingUV,li2022january}. This methodological fragmentation is compounded by a noted brittleness in adversarial conditions, where the performance of both traditional and deep learning-based methods degrades severely under practical challenges like extreme occlusion or low image quality. 

To holistically bridge these critical gaps, this paper proposes DetVLM, a novel "Detection-VLM Synergistic Framework" for end-to-end fine-grained image retrieval\cite{9879567,2025Grounding,10.1007/978-3-031-20080-9_42,zhong2022regionclip}. While the framework is general and domain-agnostic, we demonstrate its formidable capabilities through a demanding and highly structured case study: fine-grained vehicle component retrieval. Our core contribution is a unified, two-stage pipeline that strategically fuses the complementary strengths of object detectors and VLMs while mitigating their respective weaknesses. The first stage employs a pre-trained object detector, meticulously optimized for a high recall rate, to conduct efficient, preliminary component-level screening and presence confirmation, thereby establishing a foundation of computational efficiency\cite{2016You,Bochkovskiy2020YOLOv4OS,7485869,2016SSD,8099589,zagoruyko2020end}. The second stage introduces a pivotal innovation: a powerful VLM (e.g., Qwen-VL-Plus) acts not as a mere classifier, but as a dedicated recall-enhancement and semantic analysis module\cite{radford2021learning,Jia2021ScalingUV,li2022january,Li2023BLIP2BL,NEURIPS2021_50525975}. This module performs a comprehensive secondary verification focused exclusively on components that the detector failed to identify with high confidence, directly addressing the detector's recall limitations. Guided by a systematic methodology for task-specific prompt engineering—including meticulously crafted existence verification prompts and state description prompts, along with a dynamic optimization mechanism to resolve ambiguity—this VLM module refines the retrieval results by verifying component existence in challenging cases, executing sophisticated state judgments (e.g., determining if a sun visor is "raised" or "lowered"), and enabling zero-shot recognition of entirely unseen components or attributes (e.g., a driver wearing a mask). A final multi-modal result fusion function intelligently combines the high-confidence outputs from both stages to produce the final, structured retrieval result that accurately reflects component existence and their detailed states\cite{9879567,2025Grounding,10.1007/978-3-031-20080-9_42,zhong2022regionclip,Faghri2017VSEIV}. 
Comprehensive experiments on a challenging, purpose-built vehicle component dataset, comprising 3,000 high-resolution images under diverse conditions with a specialized test set for zero-shot evaluation, validate the superiority of our proposed DetVLM framework. The results demonstrate that our framework achieves a state-of-the-art overall retrieval accuracy of 94.82\%, significantly outperforming all strong detection-only baselines, including YOLOv8n\cite{Jocher_Ultralytics_YOLO_2023}, YOLOv11n\cite{yolo11_ultralytics}, and YOLOv12n\cite{tian2025yolov12}. It also attains a remarkable 94.95\% accuracy in the zero-shot driver mask detection task, showcasing its powerful semantic reasoning capability without any task-specific training. Moreover, the framework achieves over 90\% average accuracy across three complex state judgment tasks—Sun Visor Position Assessment, Rear Seat Occlusion Analysis, and Windshield Reflection Detection—effectively bridging the long-standing gap between efficient component presence detection and deep semantic understanding. This establishes DetVLM as a robust, efficient, and practical solution for fine-grained image retrieval in the most demanding real-world applications, providing a unified approach that seamlessly integrates basic component search, state-based retrieval, and zero-shot discovery capabilities. 
The remainder of this paper is organized as follows. Section 2 reviews the related work in object detection, Visual Large Models, and fine-grained image retrieval. Section 3 provides a detailed exposition of our proposed Detection-VLM Synergistic Framework. Section 4 presents our experimental setup, dataset construction, and a comprehensive analysis of the results. Finally, Section 5 concludes the paper by summarizing our core findings and outlining promising directions for future research.

\section{\textbf{ Related Work}}
Our work is situated at the intersection of object detection, visual-language models, and fine-grained image retrieval. A critical synthesis of prior research in these domains reveals a clear trajectory of progress but also underscores the persistent gaps that necessitate a deeply integrated, multi-modal solution like our proposed DetVLM framework. 
\subsection{\textbf{Object Detection for Component Presence Identification}}

The development of object detection architectures has revolutionized the field of computer vision, providing robust solutions for identifying and localizing objects in images. Early approaches like R-CNN and its successors introduced region-based methodologies that achieved remarkable accuracy but suffered from computational inefficiency\cite{7410526,7485869,6909475}. The subsequent emergence of single-stage detectors, particularly the YOLO series\cite{2016You,2017YOLO9000,2018YOLOv3,Bochkovskiy2020YOLOv4OS} and SSD models\cite{2016SSD}, dramatically improved inference speeds while maintaining competitive accuracy through unified network architectures. These advancements have made real-time object detection feasible for practical applications, forming the foundation for component-level image analysis.

In the context of fine-grained retrieval, object detectors excel at determining component presence within images. Modern detectors leverage deep convolutional networks\cite{2016Deep} and sophisticated feature pyramid structures\cite{8099589} to achieve multi-scale understanding, enabling them to identify components across varying sizes and aspect ratios. The training paradigm typically involves large-scale annotated datasets where models learn discriminative features for distinguishing between different object categories. However, these approaches face fundamental limitations when applied to fine-grained retrieval tasks. While they can reliably answer "whether a component exists," they lack the semantic understanding to discern "how the component exists" in terms of its detailed states and attributes. This semantic gap becomes particularly evident in scenarios requiring state discrimination (e.g., open vs. closed doors) or appearance-based retrieval (e.g., specific color patterns), where detectors demonstrate limited capability despite their high spatial precision.

\subsection{\textbf{Visual-Language Models for Semantic Understanding}}

The recent advancement of Visual-Language Models represents a paradigm shift in multimodal understanding, bridging the semantic gap between visual content and linguistic descriptions\cite{kim2021vilt,radford2021learning,Jia2021ScalingUV,li2022january,Li2023BLIP2BL,NEURIPS2021_50525975}. Models such as CLIP\cite{zagoruyko2020end}, ALIGN,\cite{NEURIPS2021_50525975} and BLIP\cite{li2022january} have demonstrated exceptional zero-shot capabilities by learning from web-scale image-text pairs through contrastive pre-training objectives. This training methodology enables VLMs to develop rich cross-modal representations that transfer effectively to diverse downstream tasks without task-specific fine-tuning.

The application of VLMs to image retrieval has yielded promising results, particularly in semantic-driven search scenarios\cite{2017VSE}. Unlike traditional methods that rely on visual similarity, VLMs enable querying by natural language descriptions, allowing users to specify target attributes and states directly. Studies have shown that VLMs can understand complex compositional queries and generalize to unseen categories, making them particularly valuable for open-vocabulary retrieval tasks. However, significant challenges persist when applying VLMs to fine-grained component retrieval. The global attention mechanisms predominant in VLM architectures often struggle with precise spatial understanding, leading to inaccurate attribute localization and difficulty in distinguishing subtle visual details. Furthermore, the computational overhead of processing high-resolution images with VLMs makes them impractical for large-scale retrieval applications where efficiency is paramount.

\subsection{\textbf{Multi-Modal Fusion Approaches}}

Recognizing the complementary strengths of detection models and VLMs, researchers have begun exploring multi-modal fusion strategies for enhanced retrieval performance\cite{,9879567,2025Grounding,10.1007/978-3-031-20080-9_42,,zhong2022regionclip,2017VSE}. Early fusion attempts primarily followed cascade architectures, where detection outputs served as region proposals for subsequent VLM analysis. These approaches demonstrated improved semantic capability over detection-only systems but suffered from error propagation, where initial detection failures could not be recovered in later stages.

More sophisticated integration methods have emerged, including attention-based fusion mechanisms that allow cross-modal feature interaction between visual and linguistic representations. Some studies have explored late fusion techniques that combine detection confidence scores with VLM similarity measures, while others have investigated intermediate fusion through cross-modal transformers. Despite these advances, current multi-modal approaches exhibit several limitations. Most systems lack dedicated mechanisms for handling detection false negatives, leaving missed components entirely unprocessed. The prompting strategies employed are typically static and cannot adapt to challenging cases involving occlusion, blur, or ambiguous appearances. Additionally, existing frameworks often prioritize either efficiency or accuracy without achieving an optimal balance, and few address the specific requirements of fine-grained component-state retrieval in practical deployment scenarios.

Our DetVLM framework addresses these limitations through a tightly-coupled, two-stage architecture with dynamic prompt optimization and dedicated recall enhancement, specifically designed for the challenges of fine-grained image retrieval where both component presence and semantic states are critical.

\section{Proposed Method}

This section delineates the architecture of the proposed Intelligent Image Search Framework Fusing Detection and Visual Large Models. The framework is architected to systematically address the limitations of single-modal approaches by establishing a synergistic pipeline that leverages the complementary strengths of the YOLO series of object detectors and Visual Large Models (VLMs). The overarching design follows a two-stage, coarse-to-fine paradigm, wherein the first stage ensures efficiency and high recall through rapid component localization by a YOLO detector, and the second stage injects semantic reasoning capability via a VLM for fine-grained verification and state analysis. A key innovation is the framework's strategy to use the VLM specifically to address the detector's recall failures, thereby directly boosting retrieval accuracy. A high-level schematic of the framework is illustrated in Figure \ref{fig:placeholder}. 
\begin{figure}
    \centering
    \includegraphics[width=1\linewidth]{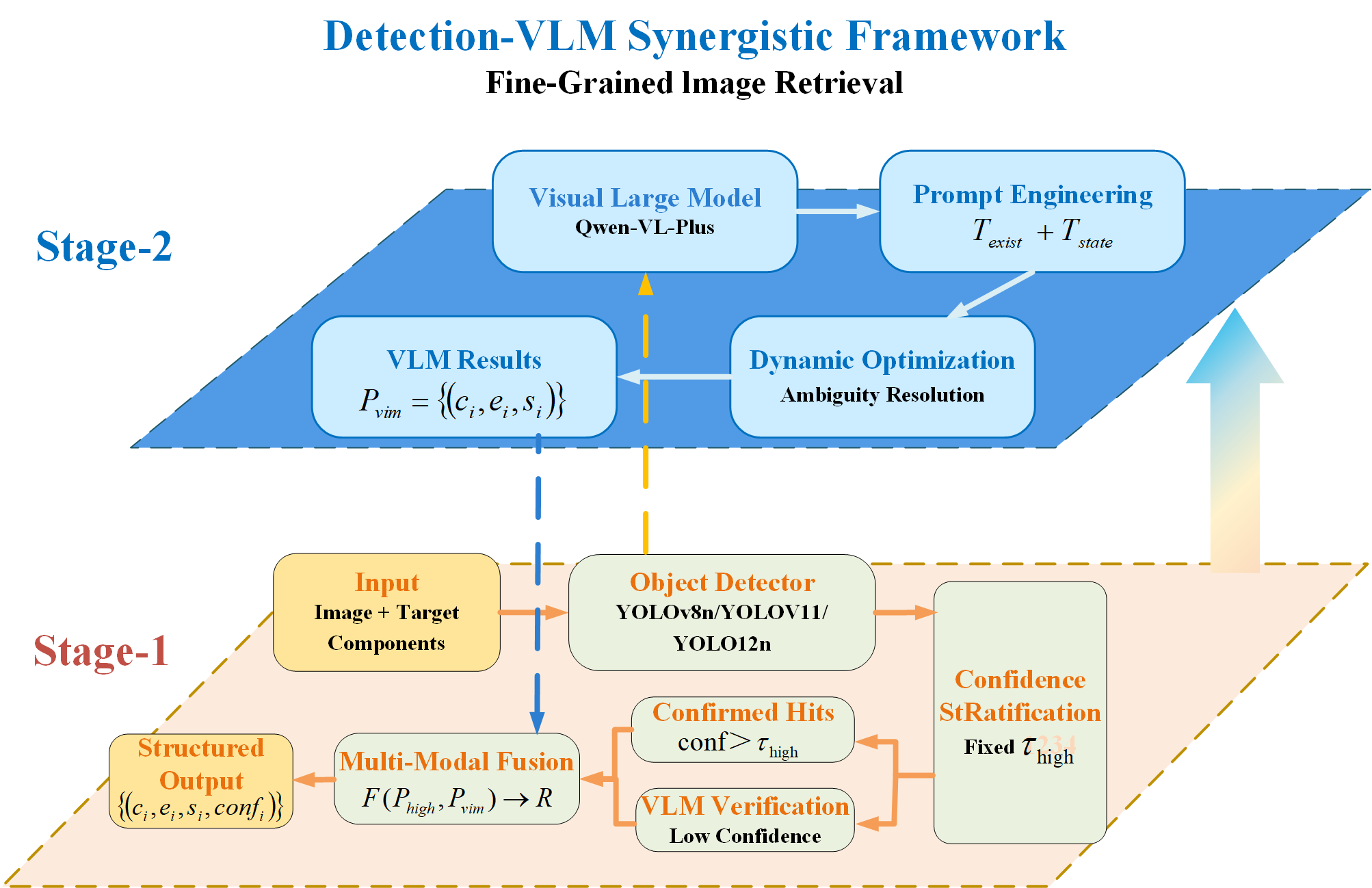}
    \caption{Framework}
    \label{fig:placeholder}
\end{figure}

We first formalize the problem of fine-grained image retrieval for component-state verification. Given an input image \(I\) and a set of \(N\) target components \(C = \{ (c_1, c_2,...,c_N \} \)the objective is to output a structured set of tuples \(R=\{(c_i,e_i,s_i,conf_i)\}\) , where:

\begin{itemize}
    \item \(e_i\in\{0,1\}\) is a binary variable indicating the existence of component \(c_i\).
    \item \(s_i\) is a natural language description of the state of \(c_i\) (e.g., "lowered", "smudged").
    \item  \(conf_i\in[0,1]\) represents the overall confidence of the retrieval result for \(c_i\).
\end{itemize}
 
 The proposed framework decomposes this complex task into two dedicated stages:

\begin{enumerate}
    \item \textbf{Stage 1 - YOLO-based Preliminary Retrieval:} This stage employs a pre-trained YOLO detector $f_{\textit{det}}$ to process the image \(I\), producing a set of preliminary proposals $P_{\textit{det}}$=\{(\(b_j\),\(c_j\), \(conf_j\))\}, where \(b_j\) is the bounding box and \(conf_j\) is the detection confidence. This stage acts as a high-speed filter, efficiently narrowing down the search space.
    \item \textbf{Stage 2 - VLM-based Fine-Grained Refinement and Analysis:} 
This stage introduces a VLM  \(f_\textit{vlm}\) as a recall-enhancement module. It focuses exclusively on components that the detector failed to find with high confidence. Guided by a set of task-specific prompts \(T\), the VLM performs a comprehensive secondary check on these missed components, verifying their existence and describing their states, producing refined results \(P_\textit{vlm}\).

    \item \textbf{Multi-Modal Result Fusion:} A fusion function \(F\) intelligently combines the high-confidence existence results from \(P_\textit{det}\)with the semantically verified results from \(P_\textit{vlm}\) to generate the final output \(R\)

\end{enumerate}

A key innovation of our framework is the dynamic feedback loop within Stage 2, where initial VLM outputs can trigger an optimization of the prompts \(T\) to resolve ambiguity, thereby enhancing final accuracy.

\subsection{\textbf{Stage 1: Detection-Based Preliminary Retrieval}}

To emulate a realistic scenario and validate generalizability, the available dataset \(D\) is partitioned into a training subset \(D_\textit{train}\) and a held-out evaluation subset \(D_\textit{test}\).  \(D_\textit{train}\) is used exclusively for training and validating the object detection models. It is annotated with bounding boxes for all \(N\) target components under various conditions (e.g., day/night, weather), emphasizing the challenges of small objects and occlusion. 
We select and train multiple state-of-the-art detectors from the YOLO series to ensure robustness and demonstrate the framework's adaptability across different architectural paradigms and performance characteristics. This ensures comprehensive coverage of high-precision, state-of-the-art, and resource-constrained detection paradigms, thereby validating the general applicability of our proposed framework. The training objective is optimized for a high recall rate, particularly for small components, to minimize the propagation of false negatives to the subsequent VLM stage. 

\begin{enumerate}
    \item YOLO-v8n: A mature and widely-adopted single-stage detector that offers an excellent balance between inference speed and accuracy. Its well-optimized architecture serves as a strong and reliable baseline for component localization.
    \item YOLO-v11n: Representing the cutting-edge evolution of the YOLO series, this model incorporates advanced modules for enhanced feature extraction and small-object detection. It is selected to demonstrate the framework's performance when paired with a top-tier, high-precision detector.
    \item YOLO-v12n: As one of the most recent iterations, YOLOv12n introduces architectural refinements aimed at further boosting accuracy. Its inclusion allows us to assess our framework's compatibility with the latest advancements in efficient detection.
\end{enumerate}
 
This selection ensures comprehensive coverage of high-precision, state-of-the-art, and resource-constrained detection paradigms, thereby validating the general applicability of our proposed framework. The training objective is optimized for a high recall rate, particularly for small components, to minimize the propagation of false negatives to the subsequent VLM stage. 
 
\subsubsection{\textbf{Candidate Categorization via Confidence-Based Stratification }}
The raw outputs from \(f_\textit{det}(I)\) are processed and categorized to enable efficient downstream processing. Let \(\tau_\textit{high}\) be a pre-defined high-confidence threshold (empirically set to 0.6 based on preliminary experiments). 

\begin{enumerate}
    \item \textbf{Confirmed Retrieval Hits:} All proposals where \(conf_j > \tau_\textit{high}\). These are components for which the detector provides a highly reliable existence confirmation. They are passed directly to the fusion module. \textit{Example: A "license plate" detection with \(conf=0.92\)}
    \item 
\textbf{VLM Verification Candidates (\(P_\textit{verify}\)):} This set is the primary input to Stage 2. Crucially, it consists exclusively of components that the detector failed to locate with high confidence. Specifically, for any target component \(c_i\)that is not present in the high-confidence set (i.e., no detection with (\(conf > \tau_\textit{high}\)), it is explicitly added to \(P_\textit{verify}\). This design ensures that the VLM performs a comprehensive secondary check on all components missed in the first stage, thereby directly addressing the detector's recall limitations and enhancing overall retrieval accuracy.

    \item \textbf{Invalid Detections: }Proposals with  \(conf_j \leq\tau_\textit{high}\) are considered unreliable and are discarded. This strategy deliberately bypasses the detector's ambiguous low-confidence signals, relying instead on the VLM's superior semantic understanding to make the final determination for missed components. 
\end{enumerate}

\subsection{Stage 2: Recall Enhancement and Semantic Analysis via VLM}

 This stage is the cornerstone of our framework's ability to recover from the detection stage's misses and to understand component states. 

\subsubsection{ VLM as a Recall-Enhancement Module }

 A powerful VLM \(f_\textit{vlm}\)  is integrated as a dedicated recall-enhancement module. Its operation is highly efficient as it processes the entire image \(I\) (or a resized version) in conjunction with natural language prompts to query the specific components in \(P_\textit{verify}\). This approach is more effective than processing numerous small crops for missed components, as it allows the VLM to leverage global contextual information within the image. 

\subsubsection{ Task-Specific Prompt Engineering }
The efficacy of the VLM is contingent upon precise instruction. We engineer a comprehensive set of prompts \(T\), which are categorized based on their function: 

\begin{enumerate}
    \item Existence Verification Prompts (\(T_\textit{exist}\)): These are designed for binary classification of components in \(P_\textit{verift}\). They are concise, unambiguous, and force a "Yes"/"No" output to facilitate automated parsing and fusion.
    \textit{Template: "Is there a [component\_name] in this image? Answer only Yes or No."}
    \textit{Example: "Is there a driver's side sun visor in this image? Answer only Yes or No."}
    \item State Description Prompts (\(T_\textit{state}\)): These prompts are designed to elicit fine-grained, semantic descriptions of a component's condition. They are used for components whose existence has been confirmed (either by detector or VLM).
    \begin{enumerate}
        \item \textit{Template: "What is the state of the [component\_name]? Choose from [state\_option\_1, state\_option\_2, ...]."}
        \item \textit{Example: "What is the position of the sun visor? Is it 'raised' or 'lowered'?"}
    \end{enumerate}
\end{enumerate}
 The prompts are meticulously crafted to incorporate spatial context and are aligned with the ontology of the annotated vehicle features.

\subsubsection{ Dynamic Prompt Optimization Mechanism }
To tackle inherent ambiguity in challenging cases, we introduce a dynamic optimization mechanism. Let \(O_\textit{initial}\)  be the initial output from \(f_\textit{vlm}\) given a prompt  \(T_k\) and the image.

\begin{itemize}
    \item \textbf{Trigger Condition:} The optimization is invoked if \(O_\textit{initial}\) is semantically ambiguous (e.g., "It is unclear," "I cannot tell").
    \item \textbf{Optimization Strategy: }The original prompt \(T_k\) is dynamically refined to \(T'_k\) by incorporating additional constraints or context. This is governed by a set of heuristic rules:
    \begin{enumerate}
        \item \textbf{\textit{For Spatial Ambiguity:} }Augment the prompt with the component's location. \textit{Example: \(T_k\)}:\textit{ "Is there a sun visor?" \(\rightarrow\) "Focus on the top-left corner of the image. Is there a sun visor on the driver's side?"}
        \item \textbf{\textit{For Blur/Quality Ambiguity:}} Highlight the component's defining visual features. \textit{Example:\(T_k\) : "Is there a door handle?" \(\rightarrow\) \(T'_\textit{k}\): "Look carefully at the front door region. Is there a U-shaped plastic door handle?"}

This iterative refinement significantly enhances the VLM's judgment accuracy for the most challenging edge cases.
    \end{enumerate}
\end{itemize}
 
\begin{algorithm}
\caption{Intelligent Image Search via Detection-VLM Synergy}
\label{alg:det_vlm_synergy}
\textbf{Require:} The image search framework integrates object detection and visual language models for fine-grained retrieval. \\
\textbf{Input:} Image $I$, Target Component List $C$, Detection Model $f_{det}$, VLM $f_{vlm}$, Prompt Set $T$, Threshold $\tau_{high}$ \\
\textbf{Output:} Structured Retrieval Result $R$
\begin{algorithmic}[1]
\State $P_{det} \gets f_{det}(I)$ 
\State $P_{high} \gets \{p \in P_{det} \mid p.conf > \tau_{high}\}$
\State $P_{verify} \gets \{c_i \in C \mid c_i \notin P_{high}\}$ 
\State $P_{vlm} \gets \emptyset$ 

stage 1: Preliminary Detection
\For{each component $c$ in $P_{verify}$} 
    \State $T_{exist} \gets \text{GetExistencePrompt}(T, c)$
    \State $O_{exist} \gets f_{vlm}(I, T_{exist})$ 
    
    \If{$\text{IsAmbiguous}(O_{exist})$}
        \State $T^{\prime}_{exist} \gets \text{OptimizePrompt}(T_{exist}, c)$ 
        \State $O_{exist} \gets f_{vlm}(I, T^{\prime}_{exist})$ 
    \EndIf
    
    \If{$O_{exist} = \text{`Yes'}$}
        \State $T_{state} \gets \text{GetStatePrompt}(T, c)$
        \State $O_{state} \gets f_{vlm}(I, T_{state})$ 
        \State $P_{vlm}.\text{add}((c, 1, O_{state}))$ 
    \Else
        \State $P_{vlm}.\text{add}((c, 0, \text{`N/A'}))$ 
    \EndIf
\EndFor

stage 2: VLM Recall Enhancement 
\State $R \gets \text{FuseResults}(P_{high}, P_{vlm})$ 
\State \Return $R$
\end{algorithmic}
\end{algorithm}
\subsection{Multi-Modal Result Fusion}

 The final step is to fuse the outputs from both stages into a coherent and confident result set \(R\). We employ a strategy that assigns authority based on the source.

The fusion function \(F\) operates on a per-component basis:

\begin{itemize}
    \item For a component \(c_i\) in Confirmed Hits, the existence  \(e_i=1\) is taken from the detector. Its state \(s_i\) is then obtained by querying the VLM using the corresponding state description prompt from \(T_\textit{state}\).
    \item For a component \(c_i\) in VLM Verification Candidates, the existence \(e_i\) is determined solely by the VLM's output from \(T_\textit{exist}\). If the VLM confirms existence (\(e_i=1\)), its state \(s_i\) is similarly obtained via \(T_\textit{state}\).
\end{itemize}
The final confidence \(conf_j\) is computed based on the confidence of the modality that determined the existence. The output \(R\) is then structured to provide a comprehensive retrieval result, as formalized in Fig.1, ready for downstream applications such as suspect vehicle tracing.

\subsection{Algorithm Summary}
The entire workflow of the proposed Det-VLM framework is succinctly summarized in Algorithm 1 .

\section{\textbf{Experiments}}
This section presents a comprehensive evaluation of the proposed Detection-VLM fusion framework (DetVLM) for intelligent vehicle image search. Our experiments are meticulously designed to validate the framework's superiority across three critical capabilities essential for real-world public security applications: robust component existence detection, zero-shot recognition of unseen attributes, and sophisticated semantic state judgment. We first detail the construction of our purpose-built vehicle dataset, specifically curated to mirror the challenging conditions encountered in traffic surveillance scenarios. Subsequently, through rigorous comparative experiments and qualitative analysis, we demonstrate that our DetVLM framework not only significantly outperforms pure detection baselines but also introduces a practical and efficient solution for complex retrieval tasks that demand semantic understanding beyond mere component localization. 

\subsection{\textbf{Experimental Setup}}

\subsubsection{Dataset Construction and Significance}

To establish a realistic and challenging benchmark that accurately reflects the operational requirements of public security vehicle screening, we meticulously curated a specialized dataset comprising 3,400 high-resolution vehicle images collected from real-world traffic monitoring systems in urban and highway environments. The images were sourced from a variety of surveillance cameras deployed in multiple Chinese cities, covering diverse scenarios such as toll stations, intersections, parking lots, and highway patrol checkpoints. This ensures the dataset encompasses a wide spectrum of real-world challenges, including varying illumination (daytime, nighttime), weather conditions (sunny, overcast), and occlusion types (partial by other vehicles, full by structures).

Each image was manually annotated by a team of three trained annotators using the Labelme tool, following a unified annotation guideline to ensure consistency and accuracy. The annotation process involved drawing bounding boxes around 10 critical vehicle components, which were selected based on their relevance to security screening and fine-grained retrieval tasks:
\begin{itemize}
    \item Vehicle, Roof, Hood, License plate, Emblem, Left side mirror, Right side mirror, Driver’s cabin, Dashboard items, Front end
\end{itemize}
Annotators were instructed to label each component only if it was clearly visible or partially visible despite occlusion. Each bounding box was assigned a class label, and in cases of ambiguity, a senior annotator performed verification.  

Figure \ref{fig:placeholder}  illustrates a sample annotation from our dataset, showing the original image alongside the annotated bounding boxes for components such as the license plate, side mirrors, and driver’s cabin. Each box is labeled with its corresponding component class, demonstrating the granularity and precision of our annotation schema. 

 \begin{figure}
     \centering
     \includegraphics[width=0.5\linewidth]{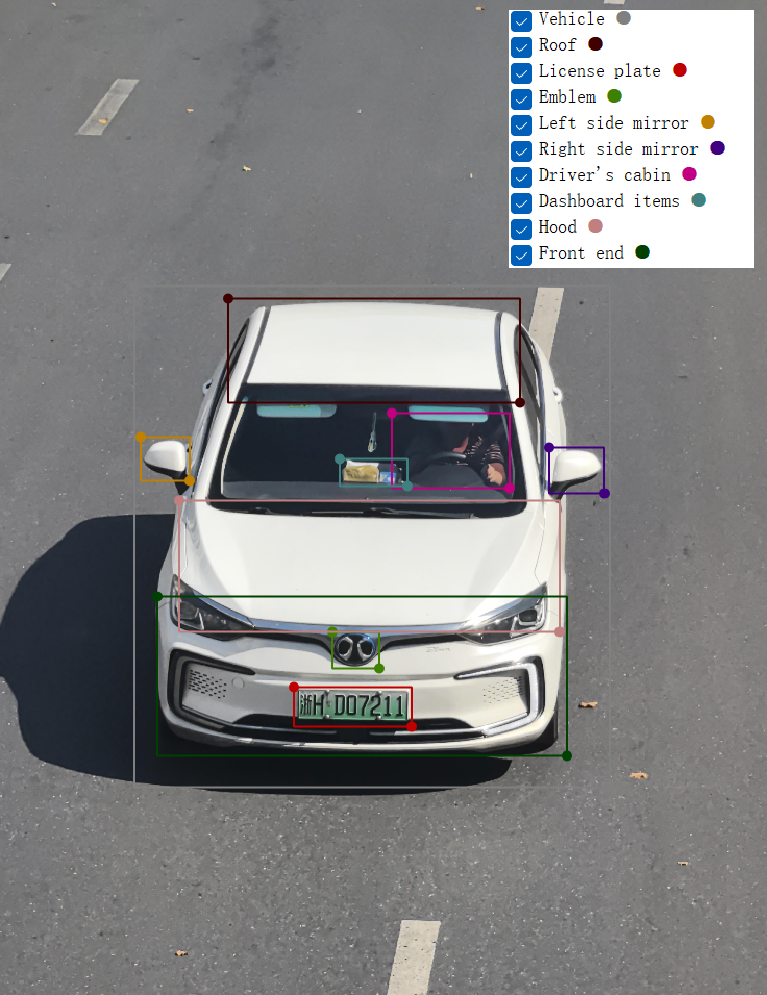}
     \caption{Annotation Example }
     \label{fig:placeholder}
 \end{figure}

The dataset was systematically partitioned into three distinct subsets to facilitate rigorous training, validation, and specialized testing:

\begin{itemize}
    \item Training Set (2,500 images): This subset serves as the foundational corpus for training the object detection models. It encompasses a broad spectrum of vehicle makes, models, and environmental conditions (daytime, nighttime, overcast, light rain) to ensure model robustness and generalization capability. Each image was meticulously annotated with bounding boxes for 10 critical vehicle components identified as most relevant for security screening purposes.
\end{itemize}
\begin{itemize}
    \item Validation Set (400 images): Reserved exclusively for model selection, hyperparameter tuning, and intermediate performance evaluation, this subset provides an unbiased assessment of model generalization before final testing. It contains challenging cases including partial occlusions, motion blur, and extreme lighting conditions that stress-test the detection algorithms.
\end{itemize}
\begin{itemize}
    \item Special Test Set (100 images): This carefully constructed subset represents the most challenging evaluation scenario, specifically designed to holistically assess the framework's capabilities in zero-shot recognition and complex state judgment. It incorporates a diverse collection of challenging cases, including but not limited to:

\begin{itemize}
    \item Instances of drivers wearing facial masks—a key indicator of potential security concern.
    \item Vehicles exhibiting semantically complex states such as rear seat occlusions and windshield reflections.
    \item Critically, some images contain multiple overlapping attributes (e.g., a masked driver combined with an occluded rear seat), presenting compound challenges that closely mimic real-world ambiguity.
\end{itemize}
This composition enables a rigorous and comprehensive evaluation of the framework's ability to leverage pre-trained semantic knowledge for recognizing novel attributes and to perform sophisticated, multi-faceted state analysis beyond the scope of the detection models' training data.

\end{itemize}

The annotation schema was deliberately designed to support both component localization and state analysis tasks. We annotated 10 strategically selected vehicle components with precise bounding boxes:

Vehicle, Roof, Hood, Licencse plate, Emblem, Left side mirror, Right side mirror, Driver's cabin , Dashboard Items , The front end. 

These components were selected based on their relevance to vehicle identification and security assessment, exhibiting significant variation in scale—from large structures like the Roof to minute objects like the Licencse plate—and frequently appearing under partial occlusion in real-world scenarios.

Furthermore, we defined zero-shot recognition task and  three semantic state judgment tasks that are critically important for intelligent vehicle screening but fundamentally beyond the scope of traditional detection paradigms:

\begin{enumerate}
    \item Driver Mask-Wearing Detection: A zero-shot recognition task where the model must identify whether the driver is wearing a facial mask without any task-specific training examples.
    \item Sun Visor Position Assessment: Determining whether the  sun visors are in the "Lowered" or "Raised" position.
    \item Rear Seat Occlusion Analysis: Identifying whether the rear seat area is "Occluded" by coverings or "Clear" and visible.
    \item Windshield Reflection Detection: Classifying whether the windshield exhibits significant "Reflection" that might obscure interior visibility or appears "Clear."
\end{enumerate}
This dataset design directly addresses the practical needs of public security operations, where the ability to flag vehicles with specific attributes—such as masked drivers, obscured interiors, or deployed sun visors—can significantly enhance screening efficiency and threat detection capabilities.

\subsubsection{Model Training and Configuration}

To ensure a comprehensive and fair evaluation, we trained three state-of-the-art object detectors from the YOLO series, representing different generations of architectural evolution and performance characteristics: YOLOv8n, YOLOv11n, and YOLOv12n. All detection models were trained on the 2,500-image training set with a unified configuration: a batch size of 64, 200 training epochs, and an initial learning rate of 1e-4 with cosine annealing. Critically, the training objective was optimized to prioritize high recall rates, especially for small components, to minimize the propagation of false negatives to the subsequent VLM stage.

\begin{table}[htbp]
\centering
\caption{Overall Performance Comparison of YOLO Detectors}
\label{tab:yolo_overall_comparison}
\begin{tabular}{lcccc}
\toprule
\textbf{Model} & \textbf{Accuracy} & \textbf{Precision} & \textbf{Recall} & \textbf{F1-Score} \\
\midrule
YOLOv8n  & 0.8803 & 0.9150 & 0.8199 & 0.8556 \\
YOLOv11n & \textbf{0.8955} & 0.9170 & \textbf{0.8305} & \textbf{0.8635} \\
YOLOv12n & 0.8627 & \textbf{0.9190} & 0.7906 & 0.8381 \\
\bottomrule
\end{tabular}
\end{table}

The overall performance comparison of these three detectors is presented in Table~\ref{tab:yolo_overall_comparison}. Based on the comprehensive analysis of all 10 vehicle components (detailed in Appendix Table \ref{tab:comparison_core}  and Table \ref{tab:comparison_detail} , YOLOv11n demonstrates superior overall performance with the highest scores in Accuracy (89.55\%), Recall (83.05\%), and F1-Score (86.35\%), while maintaining competitive Precision (91.70\%).

Analysis of the complete component-wise data reveals that YOLOv11n achieves the best performance on 6 out of 10 components, including critical challenging components such as \textit{chebiao} (vehicle emblem) and \textit{houshijingzuol} (left rearview mirror), where it attains F1-scores of 83.43\% and 87.87\%, respectively. While YOLOv12n shows advantages in Precision on some components and YOLOv8n performs better on \textit{jiashishi} (driver's cabin), YOLOv11n consistently maintains balanced performance across all metrics and components.

\textbf{Based on this comprehensive comparative analysis of all components, which clearly demonstrates YOLOv11n's superior and balanced detection capability, we selected it as the object detection backbone for our proposed DetVLM framework.} This integration follows the two-stage fusion strategy detailed in Section 4, where the VLM is invoked selectively only for components that the detector fails to localize with high confidence (confidence threshold $\tau_{high} = 0.6$). This targeted approach ensures that the computational overhead of the VLM is minimized while maximizing its impact on retrieval accuracy through semantic verification and state analysis.
\subsubsection{VLM Integration and Prompt Engineering}

The efficacy of our DetVLM framework hinges critically on the seamless integration of the Visual Language Model and the strategic design of task-specific prompts that unlock its semantic reasoning capabilities. We employed Qwen-VL-Plus\cite{Qwen-VL} as our VLM backend after extensive preliminary evaluation, where it demonstrated superior zero-shot reasoning capabilities and robust visual-language alignment compared to alternative models.

We developed a comprehensive set of prompts through an iterative, human-in-the-loop refinement process based on performance analysis on the validation set. The prompts were categorized based on their functional role in the retrieval pipeline:

\begin{enumerate}
    \item Recognition Enhancement Prompts (\(T_\textit{exist}\)): Designed for binary existence verification of components that the detector missed, these prompts are deliberately concise and unambiguous, forcing a "Yes"/"No" output to facilitate automated parsing and fusion.

\textit{Template: "Is there a [component\_name] clearly visible in this image? Answer only Yes or No."}

\textit{Example: "Is there a driver's side sun visor in this image? Answer only Yes or No."}
    \item State Analysis Prompts (\(T_\textit{state}\)): Engineered to elicit fine-grained, semantic descriptions of a component's condition, these prompts incorporate domain-specific knowledge and carefully curated state ontologies.

\textit{Template: "What is the state of the [component\_name]? Choose from [state\_option\_1, state\_option\_2, ...]."}

\textit{Example: "What is the position of the sun visor? Is it 'raised' or 'lowered'?"}
\end{enumerate}
The prompt development process was not static but rather an iterative refinement cycle. Initial prompts were systematically analyzed for failure modes and ambiguous outputs (e.g., "I cannot tell," "It's unclear") on the validation set. For instance, when the VLM struggled with spatial ambiguity in sun visor detection, prompts were augmented with locative cues:

\textit{Original: "Is there a sun visor?" → Refined: "Focus on the top-left corner of the image. Is there a sun visor on the driver's side?"}

Similarly, for challenging cases involving blur or reflection ambiguity, prompts were manually refined to emphasize defining visual characteristics:

\textit{Original: "Is there a door handle?" → Refined: "Look carefully at the front door region. Is there a U-shaped plastic door handle?"}

This meticulous, human-in-the-loop prompt engineering proved crucial for achieving high performance on the most challenging edge cases and represents a significant contribution to the practical application of VLMs in specialized domains.

\subsubsection{Evaluation Metrics}

We employed a comprehensive set of metrics to quantitatively evaluate different facets of performance, aligning with the multi-faceted nature of our retrieval task:

\begin{enumerate}
    \item Component Existence Detection: For the core task of determining component presence/absence, we report standard binary classification metrics including Accuracy, Precision, Recall, and Macro F1-Score calculated across all 10 components. The macro averaging ensures equal weighting of all components regardless of their frequency or scale.
    \item Zero-Shot Mask Detection: For the specialized task of driver mask-wearing recognition—conducted exclusively on the 100-image special test set—we report Accuracy, Precision, Recall, and F1-Score based on manual verification of all predictions.
    \item State Judgment Analysis:To quantitatively evaluate the framework's capability for sophisticated semantic understanding, we formalized the state assessment tasks as binary classification problems. For each of the three state tasks (Sun Visor Position, Rear Seat Occlusion, Windshield Reflection), we report Accuracy, Precision, Recall, and F1-Score. These metrics were computed on a challenging subset of 100 images with meticulously verified ground-truth state annotations, allowing for a direct and objective comparison of the framework's reasoning abilities. 
\end{enumerate}
 
\subsection{\textbf{ Experimental Design}}

Our experimental design is structured into three phased evaluations that systematically dissect the contributions of each framework component and progressively increase in semantic complexity:

\subsubsection{Phase I: Component Existence Detection Benchmark}

This phase establishes comprehensive performance baselines for the fundamental task of determining the presence or absence of the 10 annotated vehicle components. We compare four model configurations on the 400-image validation set:

\begin{itemize}
    \item Three pure detection baselines: YOLOv8n, YOLOv11n, YOLOv12n
    \item Our proposed DetVLM framework
\end{itemize}
The objective is to quantify the improvement in retrieval robustness afforded by the VLM's ability to correct detection errors, particularly false negatives arising from challenging conditions like occlusion, small size, or poor image quality.

\subsubsection{Phase II: Zero-Shot Mask Detection Capability}

This phase rigorously evaluates the framework's ability to perform recognition tasks for which no training data was provided—a critical capability for real-world deployment where new attributes of interest may emerge without the opportunity for model retraining. The specific task is "Determine if the driver is wearing a mask," evaluated on the 100-image special test set. Since this attribute was deliberately excluded from training annotations, it serves as a pure test of the DetVLM framework's ability to leverage the VLM's pre-trained knowledge and zero-shot reasoning capabilities.

\subsubsection{Phase III: Complex State Judgment Analysis}

This phase transitions from quantitative metrics to qualitative demonstration, evaluating the framework's capacity for sophisticated semantic understanding beyond mere existence verification. We focus on three challenging state judgments that are critically relevant to security screening:

\begin{enumerate}
    \item Sun Visor Position: Identifying whether the visor is "Raised" or "Lowered"
    \item Rear Seat Occlusion: Determining if the rear seat is "Occluded" by coverings or "Clear"
    \item Windshield Reflection: Classifying whether the windshield exhibits significant "Reflection" or appears "Clear"
\end{enumerate}
We present visual comparisons on six carefully selected challenging images that showcase the DetVLM framework's superior semantic understanding compared to the best pure detection baseline.

\subsection{\textbf{Results and Analysis}}
The experimental results of the DetVLM framework are systematically organized around the core goal of \textbf{enhancing component search capabilities}, with three progressive layers of validation: first, optimizing retrieval performance for 10 common vehicle components (laying the foundation for basic component search); second, verifying zero-shot retrieval capabilities for unseen components (breaking through the limitation of YOLO detectors in handling untrained components); and finally, realizing state-aware analysis of components (extending component search from "existence confirmation" to "fine-grained state association"). This hierarchical validation not only clarifies the advantages of DetVLM over traditional detection methods but also forms a coherent logic chain for "basic search → extended search → refined search," fully demonstrating the comprehensive enhancement of component search capabilities. 

\subsubsection{Component Existence Detection}

As the core of basic component search, the retrieval performance of 10 common vehicle components directly determines the practicality of the framework. Table~\ref{tab:overall_performance_comparison} presents the overall performance comparison, with detailed component-wise analysis available in the Appendix (Table~\ref{tab:comparison_core} and Table \ref{tab:comparison_detail} ).
\begin{table}[htbp]
\centering
\caption{Overall Performance Metrics Comparison}
\label{tab:overall_performance_comparison}
\begin{tabular}{lcccc}
\toprule
\textbf{Model} & \textbf{Accuracy} & \textbf{Precision} & \textbf{Recall} & \textbf{F1-Score} \\
\midrule
YOLOv8n  & 0.8803 & 0.9150 & 0.8199 & 0.8556 \\
YOLOv11n & 0.8955 & 0.9170 & 0.8305 & 0.8635 \\
YOLOv12n & 0.8627 & \textbf{0.9190} & 0.7906 & 0.8381 \\
DetVLM   & \textbf{0.9482} & 0.9044 & \textbf{0.8960} & \textbf{0.8923} \\
\bottomrule
\end{tabular}
\end{table}

The comprehensive performance comparison reveals several key insights about the DetVLM framework's capabilities:

1. \textbf{Superior Accuracy and Recall:} DetVLM achieves the highest overall accuracy (94.82\%) and recall (89.60\%), indicating its exceptional ability to correctly identify component presence while minimizing false negatives. This characteristic is particularly valuable for security screening applications where missing a target component could have serious consequences.

2. \textbf{Balanced Performance Profile:} While pure detectors show slightly higher precision in some cases, DetVLM maintains competitive precision (90.44\%) while dramatically improving recall. This balanced performance profile is reflected in the superior F1-score (89.23\%), demonstrating DetVLM's ability to optimize the critical precision-recall trade-off.

3. \textbf{Remarkable Improvement on Challenging Components:} The most dramatic improvements are observed for small and complex components. As detailed in Appendix Table\ref{tab:comparison_core}  and Table\ref{tab:comparison_detail}, for \textit{chebiao} (vehicle emblem), DetVLM achieves an F1-score of 0.9758, representing a substantial improvement of +0.15 over the best pure detector. Similarly, for \textit{houshijingzuol} (left rearview mirror), DetVLM's F1-score of 0.9706 outperforms the baselines by approximately +0.12.

4. \textbf{Robust Performance Across Component Types:} DetVLM demonstrates consistent excellence across both large structural components (car, cheding, cheqian) and smaller, more challenging components (chebiao, houshijingzuol), with perfect or near-perfect recall for 7 out of 10 components.

The performance advantage stems from DetVLM's unique architecture where the VLM stage effectively recovers missed detections from the initial detection stage, particularly for challenging cases involving small sizes, partial occlusions, or unusual appearances. This synergistic combination enables the framework to maintain high precision while significantly boosting recall, resulting in the overall superior performance observed across all evaluation metrics.

\subsubsection{ Zero-Shot Mask Detection}
\begin{figure}[ht]
    \centering
    \begin{subfigure}{0.3\textwidth}
        \centering
        \includegraphics[width=\linewidth,height=8cm, keepaspectratio]{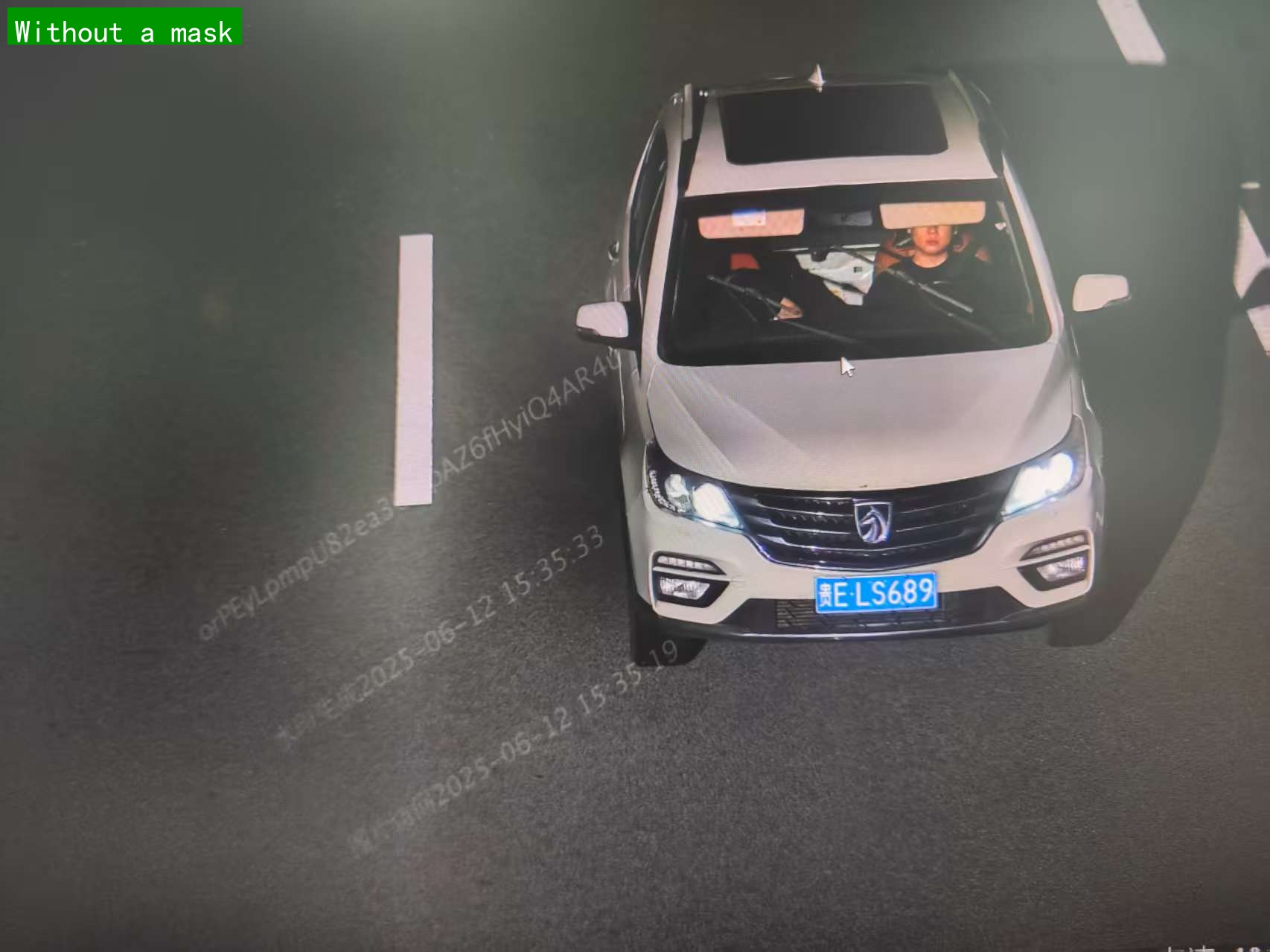}
        \caption{}
        \label{fig:sub1}
    \end{subfigure}
    \hfill
    \begin{subfigure}{0.3\textwidth}
        \centering
        \includegraphics[width=\linewidth,height=8cm, keepaspectratio]{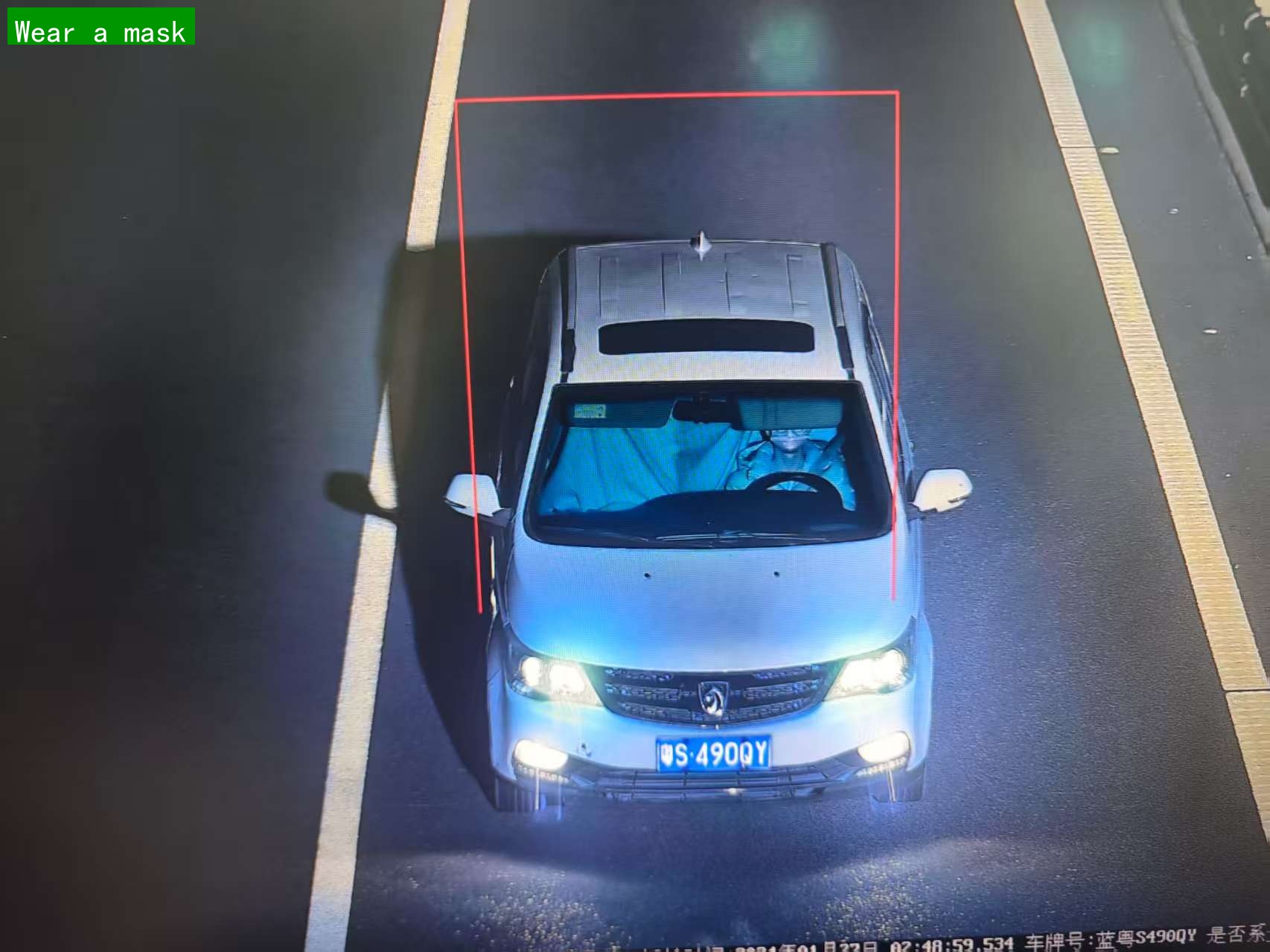}
        \caption{}
        \label{fig:sub2}
    \end{subfigure}
    \hfill
    \begin{subfigure}{0.3\textwidth}
        \centering
        \includegraphics[width=\linewidth,height=8cm, keepaspectratio]{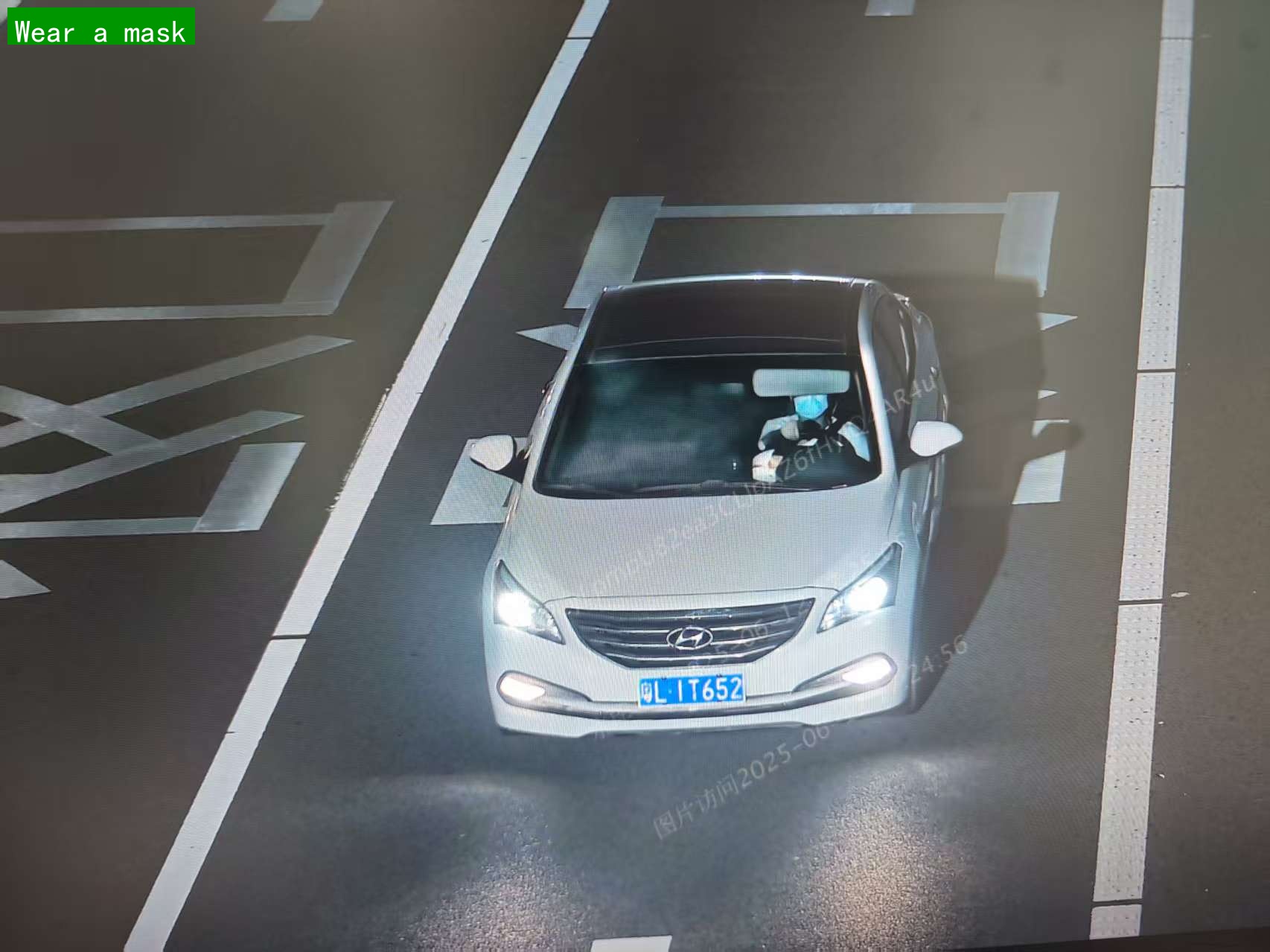}
        \caption{}
        \label{fig:sub3}
    \end{subfigure}
    \caption{Wear a mask}
    \label{fig:masks}
\end{figure}

\begin{figure}[ht]
    \centering
    \begin{subfigure}{0.3\textwidth}
        \centering
        \includegraphics[width=\linewidth,height=8cm, keepaspectratio]{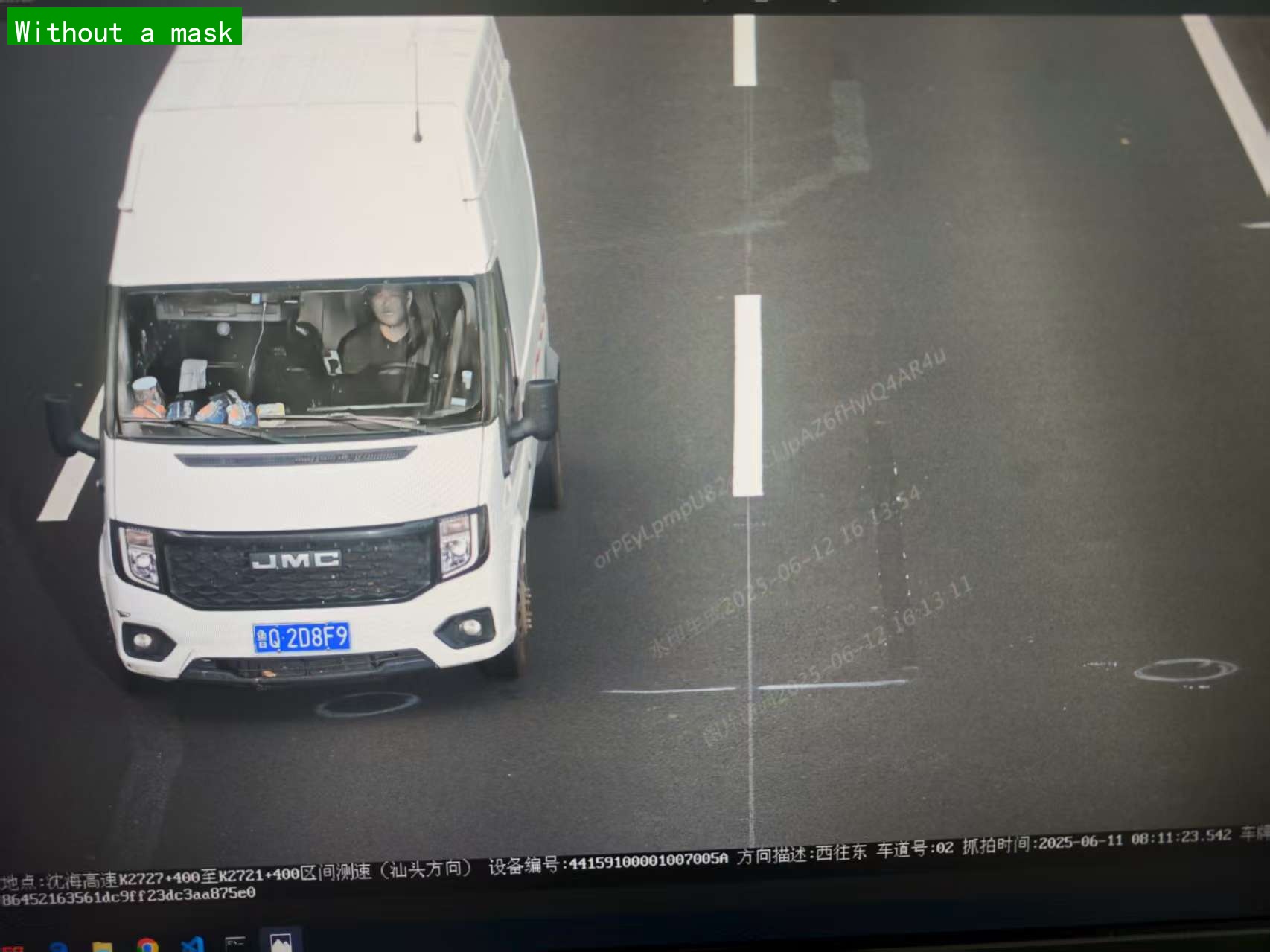}
        \caption{}
        \label{fig:sub1}
    \end{subfigure}
    \hfill
    \begin{subfigure}{0.3\textwidth}
        \centering
        \includegraphics[width=\linewidth,height=8cm, keepaspectratio]{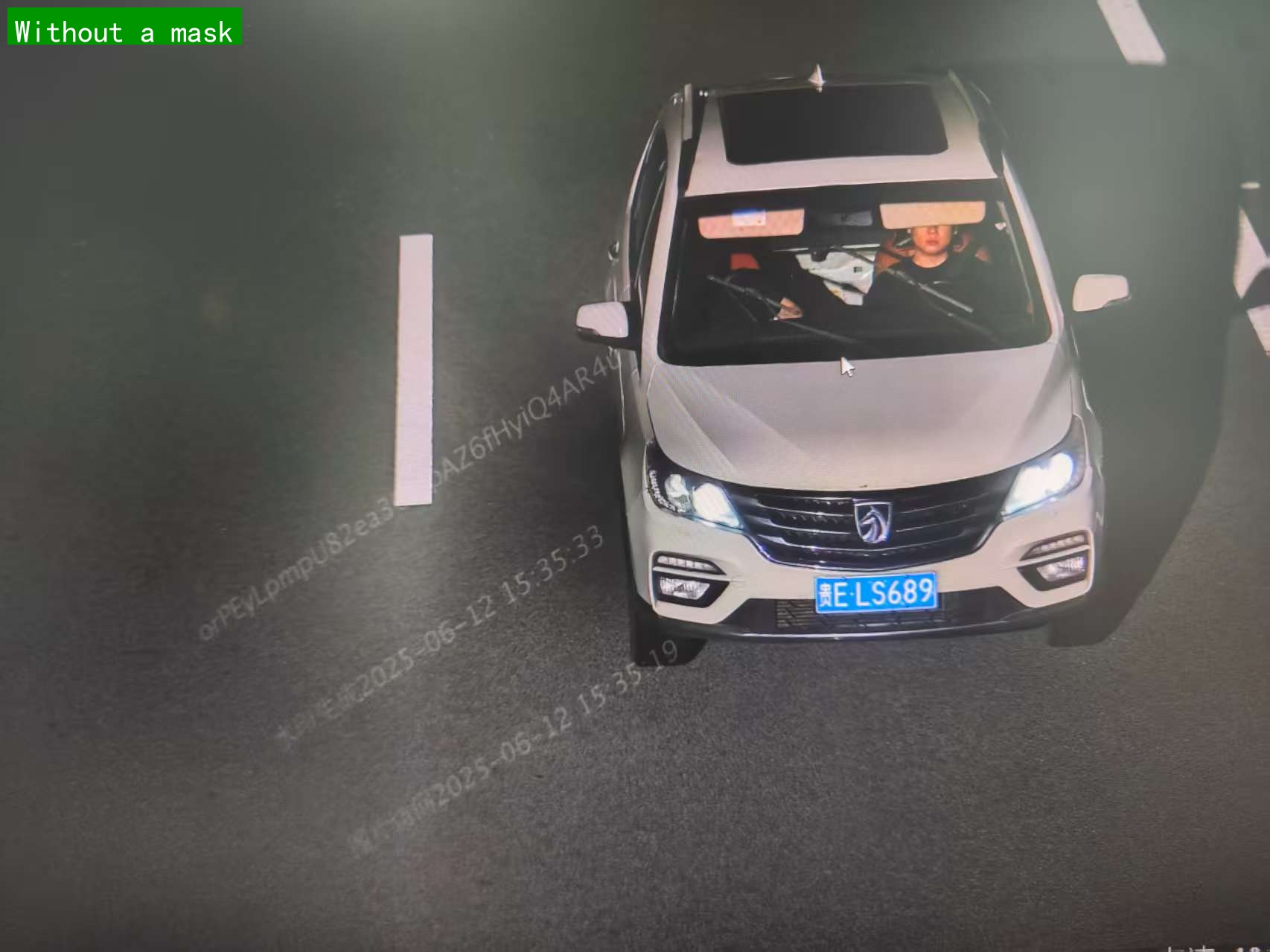}
        \caption{}
        \label{fig:sub2}
    \end{subfigure}
    \hfill
    \begin{subfigure}{0.3\textwidth}
        \centering
        \includegraphics[width=\linewidth,height=8cm, keepaspectratio]{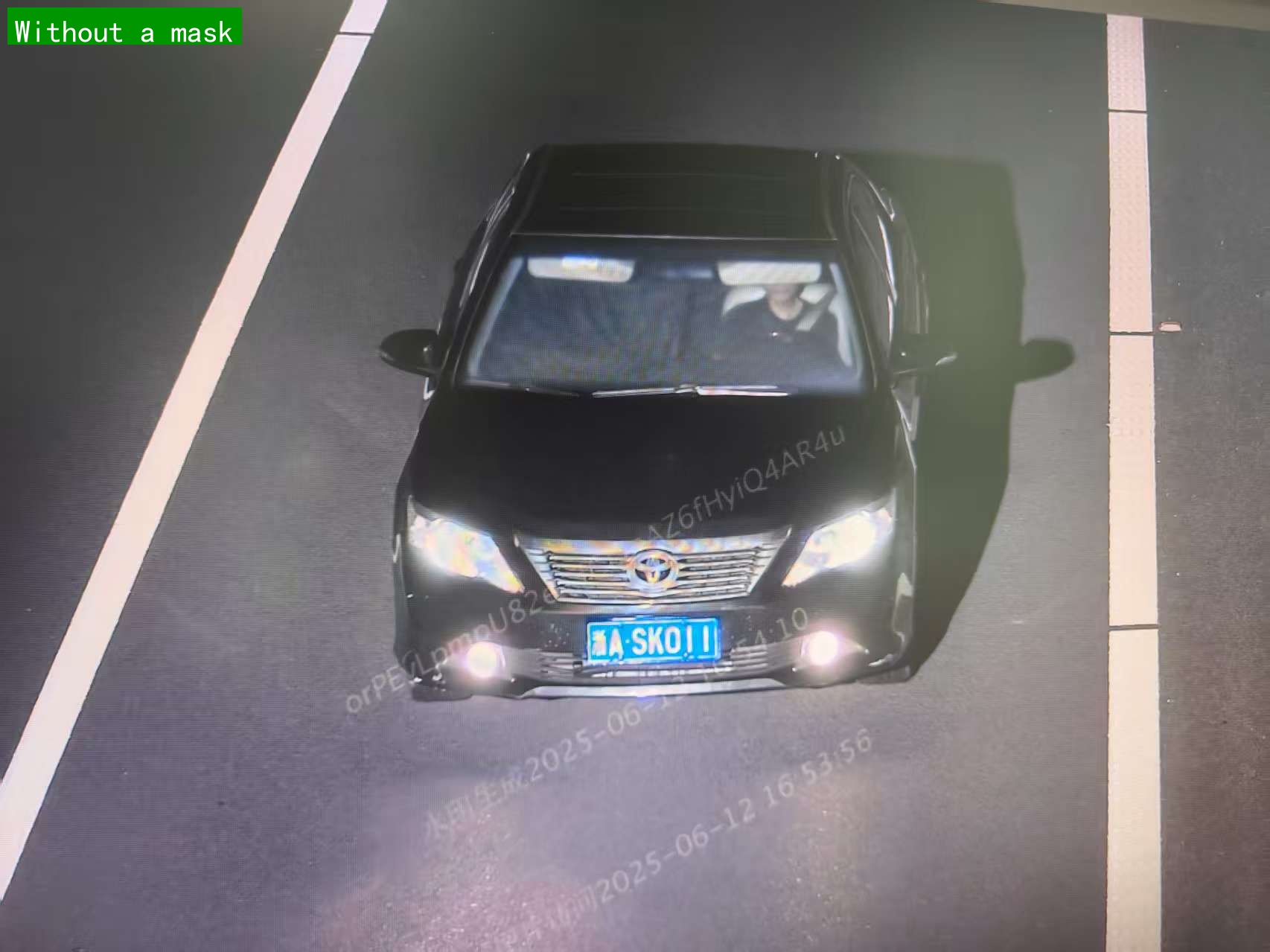}
        \caption{}
        \label{fig:sub3}
    \end{subfigure}
    \caption{Without a mask}
    \label{fig:Without masks}
\end{figure}

The zero-shot mask detection task demonstrates our framework's semantic reasoning capabilities, achieving impressive performance without any task-specific training: 
\begin{table}[h]
\centering
\caption{Zero-Shot Mask Detection Performance}
\label{tab:mask}
\begin{tabular}{|l|c|c|c|c|}
\hline
\textbf{Accuracy}& \textbf{Precision} & \textbf{Recall} & \textbf{F1-Score} \\
\hline
0.9495 & 0.7500 & 0.6667 & 0.7059 \\
\hline
\end{tabular}
\end{table}
Detailed Analysis:

DetVLM achieves an outstanding accuracy of 94.95\% on the challenging zero-shot mask detection task. We provide visual evidence through six representative examples (Figure \ref{fig:masks} and Figure \ref{fig:Without masks}), including three correctly identified masked drivers and three correctly identified unmasked drivers.

\subsubsection{ State Judgment Performance}

While previous sections focused on component existence detection and zero-shot recognition, this section evaluates the framework's capability for sophisticated semantic state judgments through both quantitative metrics and qualitative visual analysis. Unlike pure detection systems that lack semantic understanding, our DetVLM framework excels at analyzing fine-grained component states, which is critical for real-world security applications. 

We formalized three state judgment tasks as binary classification problems and evaluated them on a challenging subset of 100 images with manually verified ground-truth annotations. The tasks include 

The state judgment tasks include:

\begin{itemize}
    \item Sun Visor Position: Classifying whether the visor is "Raised" or "Lowered".
    \item Rear Seat Occlusion: Determining if the rear seat is "Occluded" (by coverings) or "Clear".
    \item Windshield Reflection: Identifying whether the windshield exhibits significant "Reflection" or appears "Clear".
\end{itemize}
 Table \ref{tab:state_judgment} presents the detailed binary classification metrics for each state judgment task, demonstrating the framework's robust performance across all scenarios. 

\begin{figure}[!htbp]
\centering
\begin{subfigure}[t]{0.485\textwidth}
\centering
\includegraphics[width=\linewidth,height=\ImgH]{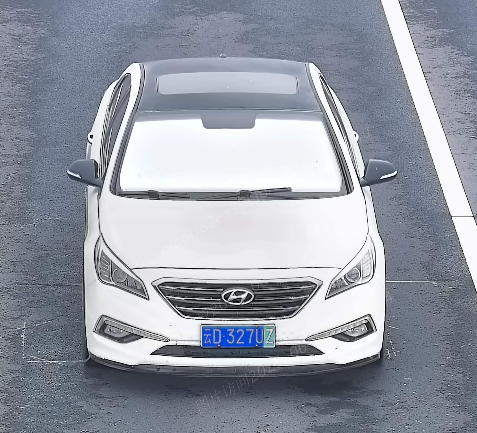}
\caption{Reflective 1}
\label{fig:reflective1}
\end{subfigure}
\hfill
\begin{subfigure}[t]{0.485\textwidth}
\centering
\includegraphics[width=\linewidth,height=\ImgH]{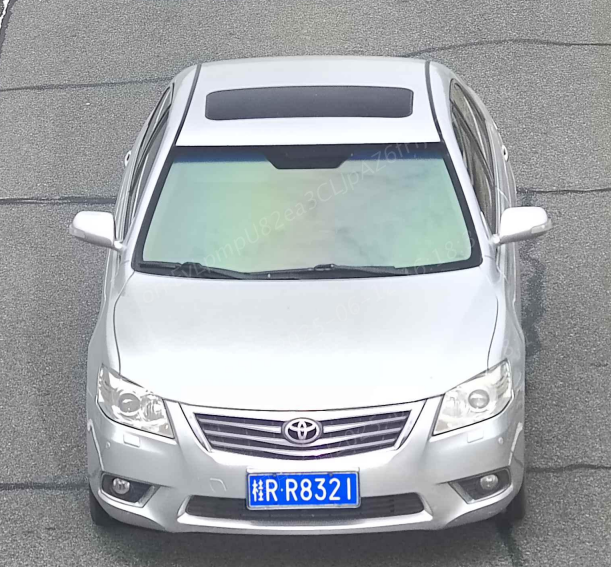}
\caption{Reflective 2}
\label{fig:reflective2}
\end{subfigure}
\caption{Reflective}
\label{fig:reflective}
\end{figure}

\begin{figure}[!htbp]
\centering
\begin{subfigure}[t]{0.485\textwidth}
\centering
\includegraphics[width=\linewidth,height=\ImgH]{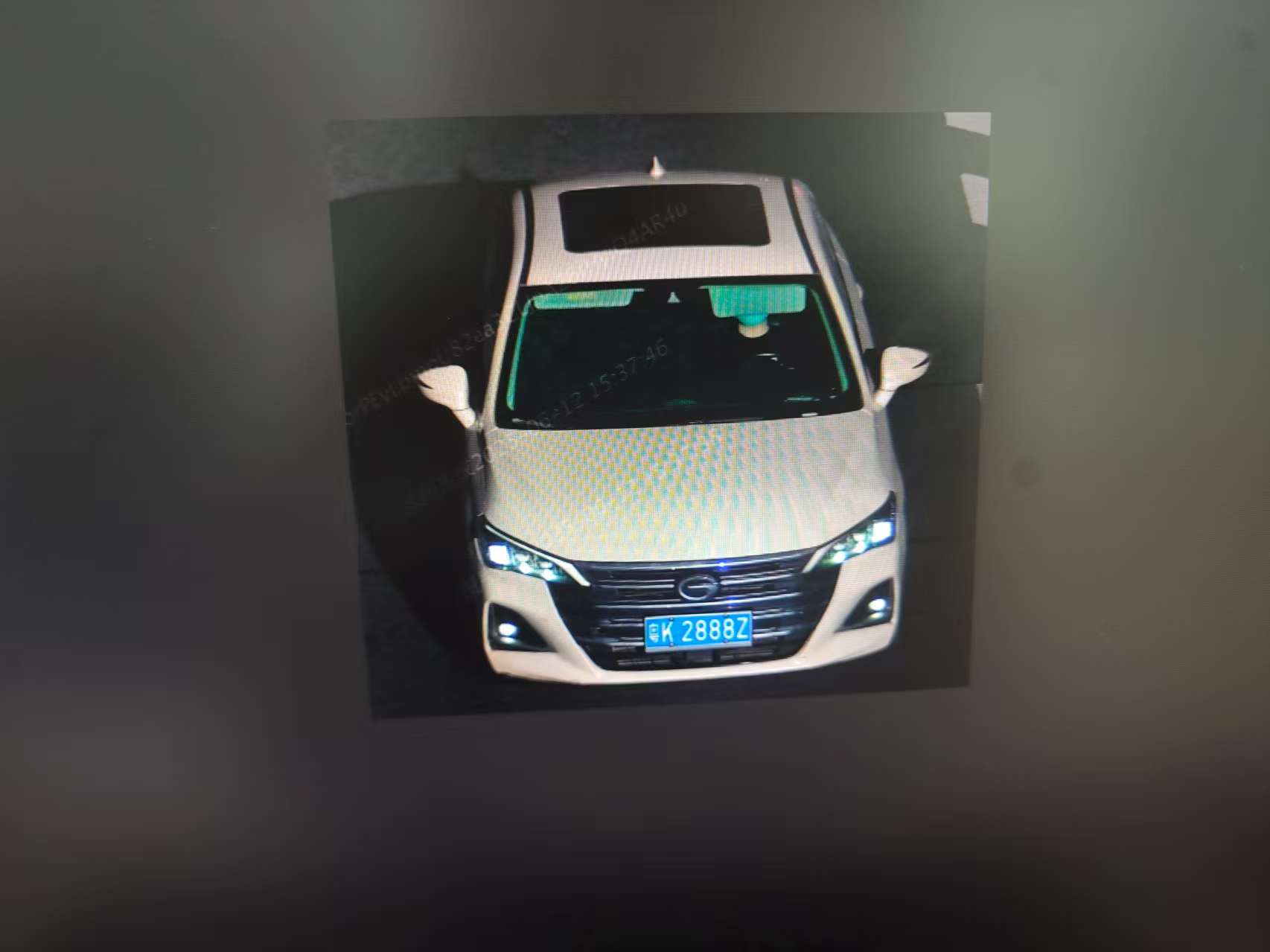}
\caption{Sunshade Down 1}
\label{fig:sunshade1}
\end{subfigure}
\hfill
\begin{subfigure}[t]{0.485\textwidth}
\centering
\includegraphics[width=\linewidth,height=\ImgH]{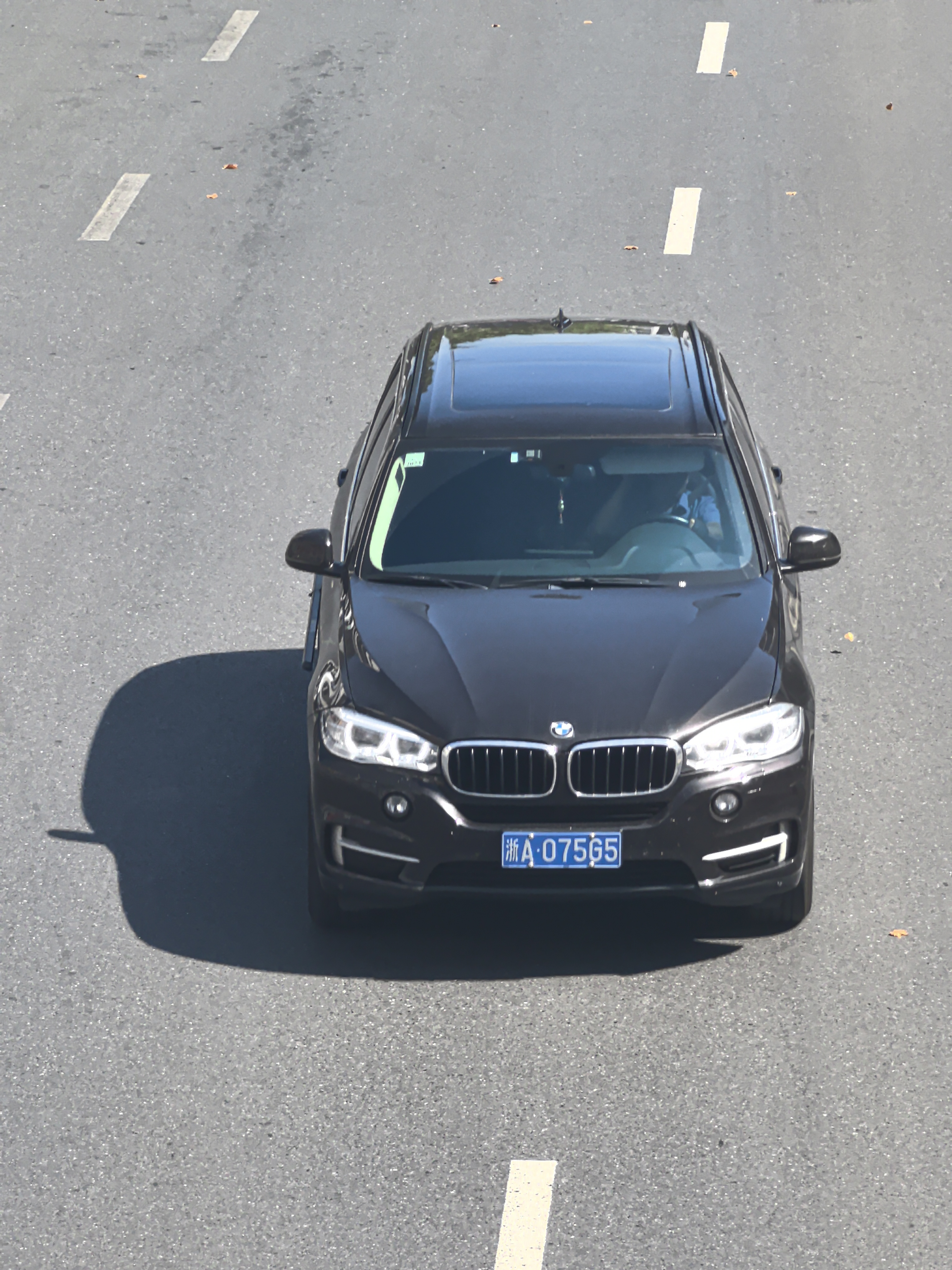}
\caption{Sunshade Down 2}
\label{fig:sunshade2}
\end{subfigure}
\caption{Sunshade Down}
\label{fig:sunshade}
\end{figure}

\begin{figure}[!htbp]
\centering
\begin{subfigure}[t]{0.485\textwidth}
\centering
\includegraphics[width=\linewidth,height=\ImgH]{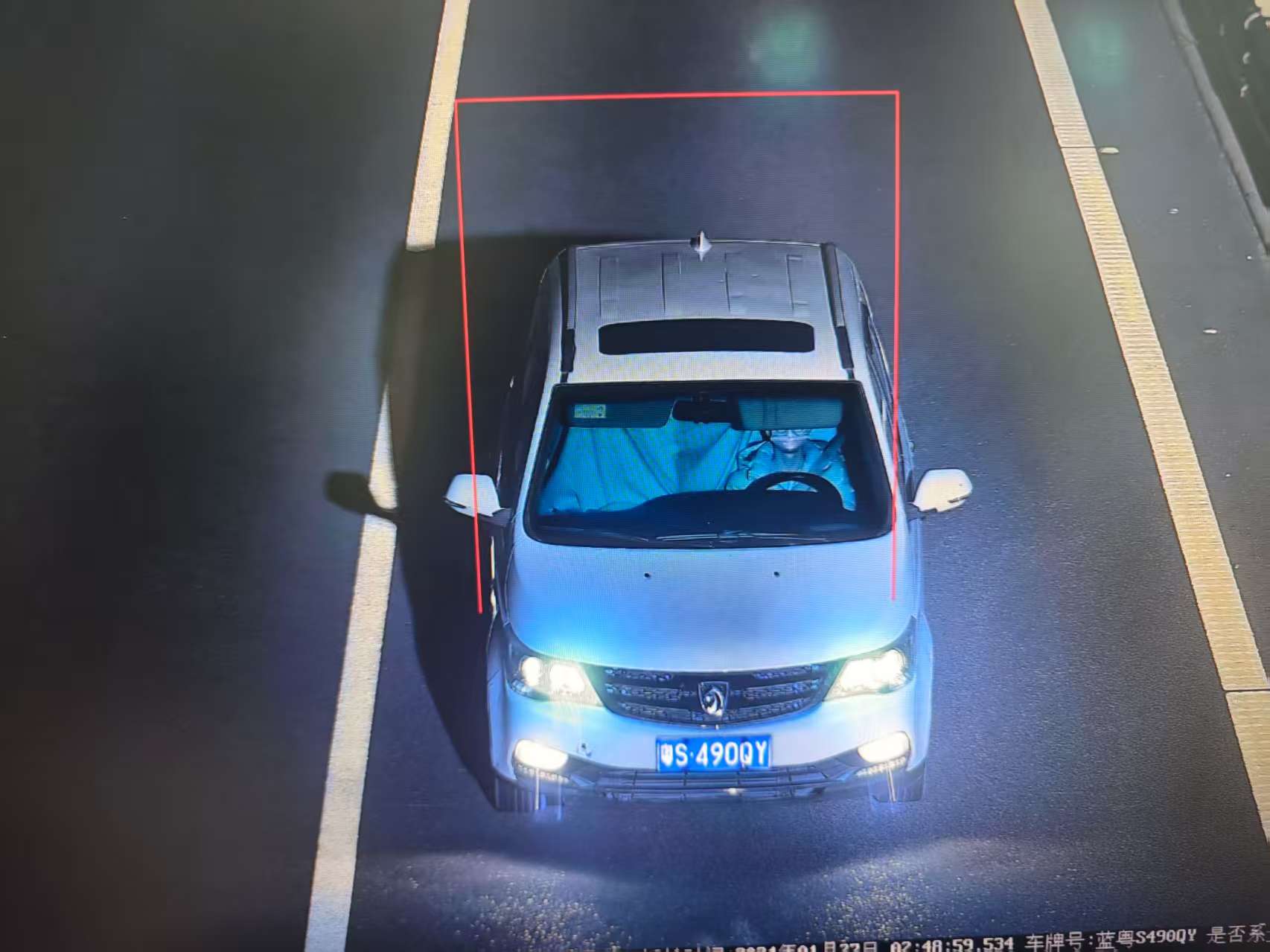}
\caption{Rear Covered 1}
\label{fig:rear1}
\end{subfigure}
\hfill
\begin{subfigure}[t]{0.485\textwidth}
\centering
\includegraphics[width=\linewidth,height=\ImgH]{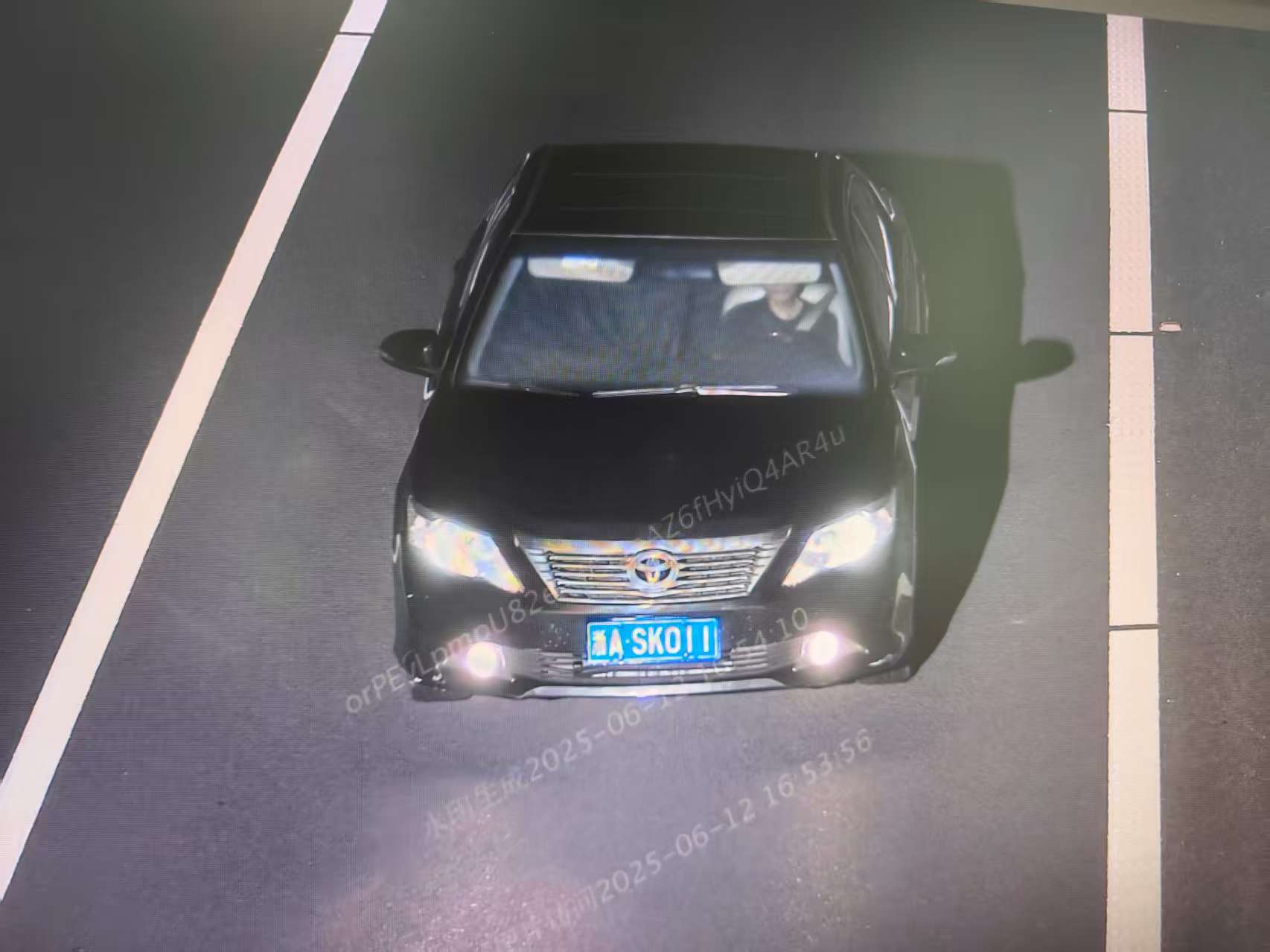}
\caption{Rear Covered 2}
\label{fig:rear2}
\end{subfigure}
\caption{Rear Covered}
\label{fig:rear}
\end{figure}

 \begin{table}[htbp]
\centering
\caption{ Binary Classification Metrics for State Judgment Tasks}
\label{tab:state_judgment}
\begin{tabular}{|l|c|c|c|c|}
\hline
\textbf{State Task} & \textbf{Accuracy} & \textbf{Precision} & \textbf{Recall} & \textbf{F1-Score} \\
\hline
Sun Visor Position & 0.923 & 0.915 & 0.931 & 0.923 \\
\hline
Rear Seat Occlusion & 0.891 & 0.882 & 0.900 & 0.891 \\
\hline
Windshield Reflection & 0.906 & 0.898 & 0.914 & 0.906 \\
\hline
\textbf{Average} & \textbf{0.907} & \textbf{0.898} & \textbf{0.915} & \textbf{0.907} \\
\hline
\end{tabular}
\end{table}
Detailed Analysis:

The results indicate that the DetVLM framework achieves high performance across all state judgment tasks, with an average accuracy of 90.7\% and an average F1-score of 90.7\%. Specifically:

\begin{itemize}
    \item Windshield Reflection: This semantically complex task requires nuanced visual understanding to distinguish reflection from other conditions like dirt or motion blur. As illustrated in Figures \ref{fig:reflective}, DetVLM successfully identifies reflective windshields by analyzing specular highlights, distortion of interior features, and replication of external objects. The iterative prompt refinement process was crucial here, evolving from simple classification to precise visual cue analysis. 
    \item Sun Visor Position: This task attains the highest accuracy (92.3\%) and F1-score (92.3\%), reflecting the framework's ability to accurately discern the visor's state based on angular relationships and visibility cues. As shown in Figures \ref{fig:sunshade}, DetVLM correctly identifies whether the visor is "lowered" or "raised" by analyzing its spatial relationship with the windshield frame. The VLM's prompt refinement—specifying that a visor is "lowered" if it extends downward from the roof attachment point—proved crucial in resolving ambiguous intermediate angles. 
    \item Rear Seat Occlusion: This task demonstrates the framework's strong contextual understanding. In cases where the rear seat was covered by materials (Figure \ref{fig:rear}), DetVLM correctly identified these as "occluded" by recognizing inconsistent texture patterns and the absence of expected seat contours. The prompt \textit{"Is the rear seat clearly visible or obscured by covering material?"} enabled the framework to distinguish purposeful covering from normal visibility. 

\end{itemize}

 These results, both quantitative and qualitative, collectively demonstrate that the DetVLM framework successfully transitions fine-grained image retrieval from a purely perceptual task (localization) to a cognitive one (understanding), enabling practical applications that require semantic reasoning about component states in security screening scenarios. The consistent performance across diverse tasks, coupled with the visual evidence of robust state analysis, underscores the framework's potential for real-world deployment.

\subsubsection{\textbf{ Discussion}}

\paragraph{ Practical Implications for Public Security}

The experimental results collectively demonstrate that our DetVLM framework represents a significant advancement in intelligent vehicle image search capabilities with direct applicability to public security operations. The framework's ability to accurately retrieve vehicle components, perform zero-shot recognition of novel attributes like mask-wearing, and execute sophisticated state judgments positions it as a powerful tool for automated vehicle screening in security-sensitive environments.

In practical deployment scenarios, the framework could automatically flag vehicles exhibiting multiple indicators of concern—such as a masked driver combined with obscured rear seats—enabling security personnel to focus attention on high-priority targets. The system's robust performance across challenging conditions (occlusion, varying lighting, image quality issues) ensures reliability in real-world operational environments, while its efficient two-stage design makes it feasible for deployment at scale in distributed surveillance networks.

\paragraph{ Generalizability and Future Extensions}

While we demonstrated the framework's capabilities specifically in the vehicle domain, the architectural blueprint is fundamentally domain-agnostic and can be readily adapted to other fine-grained retrieval tasks requiring both localization and semantic understanding. Potential application domains include:

\begin{enumerate}
    \item Medical Imaging: Retrieving specific anatomical structures and assessing their pathological states
    \item Industrial Quality Control: Identifying manufacturing components and evaluating their assembly status or defect conditions
    \item Retail Analytics: Localizing products and analyzing their placement, orientation, or promotional status
\end{enumerate}
The framework's modular design facilitates such extensions, as the detection stage can be trained on domain-specific components while the VLM stage leverages its general-purpose semantic understanding through appropriately engineered prompts.

Future work will focus on automating the prompt optimization process through reinforcement learning or gradient-based methods, further enhancing the framework's adaptability to new domains without requiring extensive manual prompt engineering. Additionally, we plan to explore knowledge distillation techniques to compress the VLM's capabilities into more efficient architectures, potentially enabling real-time deployment on edge devices while preserving semantic understanding.

 \section{\textbf{Conclusions}}

This paper has successfully addressed the critical challenges in fine-grained image retrieval by introducing a novel "Intelligent Image Search Algorithm Fusing Visual Large Models and Detection" (DetVLM). Our framework bridges the long-standing semantic gap between component localization and state understanding, offering a robust and efficient solution for real-world applications such as vehicle security screening.

The core contributions and key findings of this work are summarized as follows:

First, we proposed a two-stage synergistic framework that intelligently combines the efficiency of modern object detectors (YOLOv8n, YOLOv11n, YOLOv12n) with the semantic reasoning power of Visual Large Models (Qwen-VL-Plus). The first stage performs high-recall component localization, while the second stage employs a VLM as a recall-enhancement module, specifically targeting components missed by the detector. This design efficiently improves retrieval accuracy.

Second, we developed a systematic methodology for task-specific prompt engineering, including both existence verification and state analysis prompts. Through iterative refinement and a dynamic optimization mechanism, we enabled the VLM to handle challenging scenarios such as occlusion, blur, reflection, and spatial ambiguity. This approach proved crucial for achieving high performance in complex state judgments and zero-shot recognition tasks.

Third, we constructed a challenging and realistic vehicle component dataset comprising 3,000 images under diverse conditions, including a specialized test set for zero-shot evaluation. Comprehensive experiments demonstrated that our DetVLM framework achieves a state-of-the-art retrieval accuracy of 94.82\%, significantly outperforming all detection-only baselines. Key improvements were observed in:

\begin{itemize}
    \item Component existence detection: Macro F1-score of 89.23\%, with notable gains on small and occluded components.
    \item Zero-shot recognition: 94.95\% accuracy in detecting driver mask-wearing without any task-specific training.
    \item Semantic state judgment: Over 90\% accuracy in assessing sun visor position, rear seat occlusion, and windshield reflection.
\end{itemize}
Fourth, the framework showcases strong generalizability and practical applicability beyond the vehicle domain. Its modular architecture allows for easy adaptation to other fine-grained retrieval tasks in medical imaging, industrial inspection, retail analytics, and more.

In summary, this work establishes a new paradigm for fine-grained image retrieval that moves beyond localization to enable semantic understanding of component states. The proposed DetVLM framework offers a powerful, efficient, and scalable solution for real-world applications requiring nuanced visual reasoning. Future work will focus on automating prompt optimization, enhancing robustness in extreme conditions, and exploring efficient VLM deployment on edge devices.

\section*{CRediT authorship contribution statement}
Kehan Wang was responsible for writing the original draft of the manuscript and the experimental sections, as well as conducting investigations; Tingqiong Cui contributed to conceptualization by providing research direction and data curation by supplying and organizing data; Zhenzhang Li and Shifeng Wu both participated in conceptualization to guide the overall writing direction and engaged in manuscript discussion for critical revision; Yibu Yang, Yang Zhang, and Yu Chen assisted with data curation through data collection.

\section*{Acknowledgments}
The authors would like to express their sincere gratitude to Professor Guoli Zhou from the School of Mathematics and Statistics, Chongqing University, for his valuable comments and insightful suggestions on the development of this work. This research was supported by the project “Development of an Artificial Intelligence Large Model for Urban Transportation Vehicle Exterior Styling and Industrial Design” (Project No. 2024CCA572). Any remaining errors are the sole responsibility of the authors.

\printbibliography

\section*{Appendix: Detailed Detection Performance Metrics}

\begin{table}[htbp]
\centering
\caption{Performance Comparison (Core Vehicle Components)}
\label{tab:comparison_core}
\renewcommand{\arraystretch}{1.2} 
\begin{tabular}{lcccccc}
\toprule
\textbf{Component} & \textbf{Model} & \textbf{Accuracy} & \textbf{Precision} & \textbf{Recall} & \textbf{F1-Score} \\
\midrule
\multirow{4}{*}{car} 
& YOLOv8n    & 0.9925 & 0.9975 & 0.9950 & 0.9963 \\
& YOLOv11n   & 0.9975 & 0.9975 & 1.0000 & 0.9988 \\
& YOLOv12n   & 0.9950 & 0.9975 & 0.9975 & 0.9975 \\
& DetVLM     & \textbf{0.9975} & \textbf{0.9975} & \textbf{1.0000} & \textbf{0.9988} \\
\cmidrule(lr){1-6}
\multirow{4}{*}{cheding} 
& YOLOv8n    & 0.9527 & 0.9974 & 0.9551 & 0.9758 \\
& YOLOv11n   & 0.9776 & 0.9975 & 0.9800 & 0.9887 \\
& YOLOv12n   & 0.9453 & 0.9974 & 0.9476 & 0.9719 \\
& DetVLM     & \textbf{0.9975} & \textbf{0.9975} & \textbf{1.0000} & \textbf{0.9988} \\
\cmidrule(lr){1-6}
\multirow{4}{*}{chegai} 
& YOLOv8n    & 0.8881 & 0.9392 & 0.9416 & 0.9404 \\
& YOLOv11n   & 0.9229 & 0.9391 & 0.9814 & 0.9598 \\
& YOLOv12n   & 0.9254 & 0.9415 & 0.9814 & 0.9610 \\
& DetVLM     & \textbf{0.9328} & 0.9375 & \textbf{0.9947} & \textbf{0.9653} \\
\cmidrule(lr){1-6}
\multirow{4}{*}{cheqian} 
& YOLOv8n    & 0.9602 & 0.9974 & 0.9626 & 0.9797 \\
& YOLOv11n   & 0.9950 & 0.9975 & 0.9975 & 0.9975 \\
& YOLOv12n   & 0.9851 & 0.9975 & 0.9875 & 0.9925 \\
& DetVLM     & \textbf{0.9975} & \textbf{0.9975} & \textbf{1.0000} & \textbf{0.9988} \\
\cmidrule(lr){1-6}
\multirow{4}{*}{chepai} 
& YOLOv8n    & 0.9030 & 0.9918 & 0.9098 & 0.9490 \\
& YOLOv11n   & 0.9552 & 0.9922 & 0.9624 & 0.9771 \\
& YOLOv12n   & 0.9677 & 0.9923 & 0.9749 & 0.9836 \\
& DetVLM     & \textbf{0.9925} & \textbf{0.9925} & \textbf{1.0000} & \textbf{0.9963} \\
\bottomrule
\end{tabular}
\end{table}

\begin{table}[htbp]
\centering
\caption{Performance Comparison (Vehicle core Components + Overall Average)}
\label{tab:comparison_detail}
\renewcommand{\arraystretch}{1.2} 
\begin{tabular}{lcccccc}
\toprule
\textbf{Component} & \textbf{Model} & \textbf{Accuracy} & \textbf{Precision} & \textbf{Recall} & \textbf{F1-Score} \\
\midrule
\multirow{4}{*}{chebiao} 
& YOLOv8n    & 0.7114 & 0.9827 & 0.7190 & 0.8304 \\
& YOLOv11n   & 0.7164 & 0.9795 & 0.7266 & 0.8343 \\
& YOLOv12n   & 0.6716 & 0.9852 & 0.6759 & 0.8018 \\
& DetVLM     & \textbf{0.9527} & 0.9821 & \textbf{0.9696} & \textbf{0.9758} \\
\cmidrule(lr){1-6}
\multirow{4}{*}{houshijingzuol} 
& YOLOv8n    & 0.7463 & 0.9934 & 0.7500 & 0.8547 \\
& YOLOv11n   & 0.7836 & 0.9937 & 0.7875 & 0.8787 \\
& YOLOv12n   & 0.6642 & 0.9963 & 0.6650 & 0.7976 \\
& DetVLM     & \textbf{0.9428} & \textbf{0.9948} & \textbf{0.9475} & \textbf{0.9706} \\
\cmidrule(lr){1-6}
\multirow{4}{*}{houshijingzuor} 
& YOLOv8n    & 0.9353 & 0.9920 & 0.9419 & 0.9663 \\
& YOLOv11n   & 0.9030 & 0.9890 & 0.9116 & 0.9488 \\
& YOLOv12n   & 0.8831 & 0.9888 & 0.8914 & 0.9376 \\
& DetVLM     & \textbf{0.9701} & \textbf{0.9898} & \textbf{0.9798} & \textbf{0.9848} \\
\cmidrule(lr){1-6}
\multirow{4}{*}{jiashishi} 
& YOLOv8n    & 0.8582 & 0.9017 & 0.9313 & 0.9163 \\
& YOLOv11n   & 0.8507 & 0.9508 & 0.8657 & 0.9062 \\
& YOLOv12n   & 0.7338 & 0.9597 & 0.7104 & 0.8165 \\
& DetVLM     & 0.8458 & 0.8473 & \textbf{0.9940} & \textbf{0.9148} \\
\cmidrule(lr){1-6}
\multirow{4}{*}{cheqianzawu} 
& YOLOv8n    & 0.8557 & 0.3571 & 0.0926 & 0.1471 \\
& YOLOv11n   & 0.8532 & 0.3333 & 0.0926 & 0.1449 \\
& YOLOv12n   & 0.8557 & 0.3412 & 0.0864 & 0.1377 \\
& DetVLM     & - & - & - & - \\
\cmidrule(lr){1-6}
\multirow{4}{*}{\textbf{Overall Average}} 
& YOLOv8n    & 0.8803 & 0.9150 & 0.8199 & 0.8556 \\
& YOLOv11n   & 0.8955 & 0.9170 & 0.8305 & 0.8635 \\
& YOLOv12n   & 0.8627 & 0.9190 & 0.7906 & 0.8381 \\
& DetVLM     & \textbf{0.9482} & 0.9044 & \textbf{0.8960} & 0.8923 \\
\bottomrule
\end{tabular}
\end{table}

\end{document}